\documentclass[10pt,twocolumn,letterpaper]{article}

\usepackage{cvpr}              

\usepackage{tabularx}
\usepackage{colortbl}
\usepackage{sepfootnotes}
\usepackage{multirow}

\definecolor{cvprblue}{rgb}{0.21,0.49,0.74}
\usepackage[pagebackref,breaklinks,colorlinks,allcolors=cvprblue]{hyperref}

\definecolor{contextpurple}{RGB}{112,48,160}
\definecolor{predcyan}{RGB}{0,176,240}
\definecolor{gtyellow}{RGB}{255,192,0}

\DeclareRobustCommand{\colorboxmark}[1]{%
    \begingroup
    \setlength{\fboxsep}{0pt}%
    \setlength{\fboxrule}{1.2pt}%
    \raisebox{0.05ex}{\fcolorbox{#1}{white}{\rule{0pt}{0.8ex}\rule{0.8ex}{0pt}}}%
    \endgroup
}

\def\confName{CVPR}
\def\confYear{2027}

\newcommand{\method}{ChronoLatent\xspace}

\title{\method: Compact Dynamic Scene Latent for Novel View Synthesis}

\author{
    Zhiyuan Fang$^{1,2}$  
    \quad 
    Guo Chen$^{2,3}$ 
    \quad
    Jiarun Liu$^{2}$ 
    \quad
    Qifeng Chen$^{2}$ \\
    \quad
    Qiaowei Miao$^{1,2}$ 
    \quad
    Rengan Xie$^{1}$ 
    \quad
    Yuchi Huo$^{1}$ 
    \quad
    Sheng Yang$^{2, \dagger}$
    \\ \and
    {\normalsize $^1$Zhejiang University \quad $^2$NIO, Inc. \quad $^3$Peking University}
}

\begin{document}

\twocolumn[{%
  \renewcommand\twocolumn[1][]{#1}
  \maketitle

  \begin{center}
    \captionsetup{type=figure}
    \includegraphics[width=\textwidth]{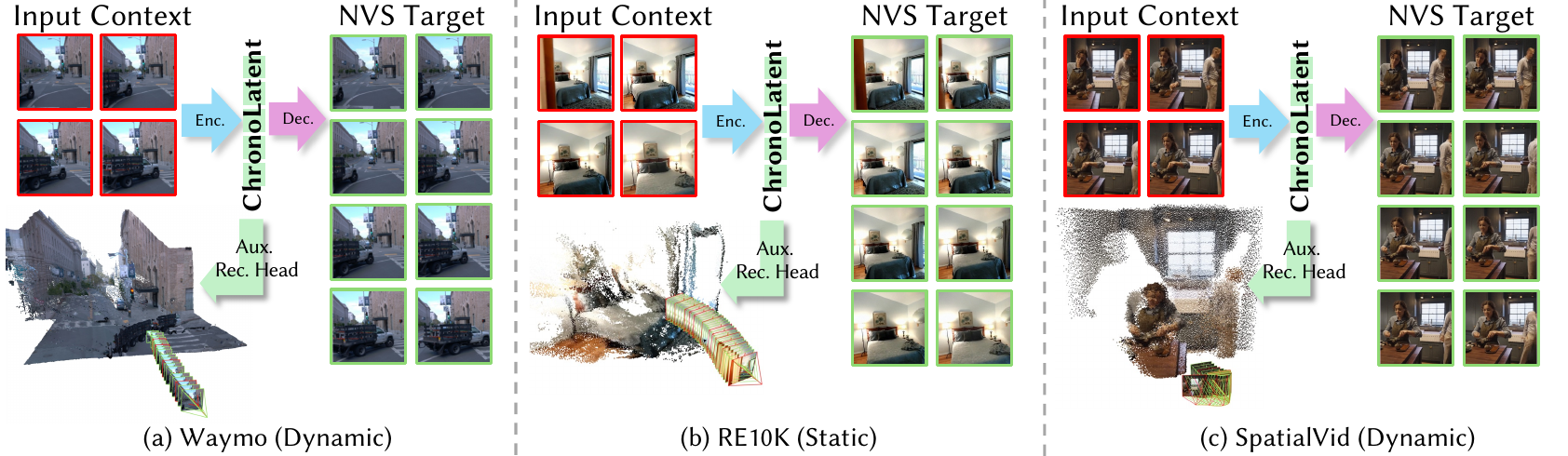}
    \captionof{figure}{\method encodes observations into a compact dynamic scene latent. Given this representation and a camera pose query and a timestamp, it generates novel views with scene-guided DiT decoder. \method achieves high-quality NVS across diverse inputs: (a) dynamic outdoor driving scenes, (b) static indoor scenes, and (c) general dynamic scenes.}
    \label{fig:teaser}
  \end{center}
}]

\renewcommand{\thefootnote}{}
\footnote{$\dagger$: corresponding author.}

\begin{abstract}
Dynamic novel view synthesis (NVS) from sparse posed RGB observations requires integrating geometry, appearance, and motion across views and time.
Although recent methods consolidate these cues into compact scene representations, faithfully synthesizing time-varying scene content remains challenging.
We present \method, a generative framework combining compact spatio-temporal scene encoding with pixel-space image synthesis.
The encoder uses Temporal-Projective RoPE (TP-RoPE) to incorporate spatial, temporal, and camera context into geometry and appearance features.
Learnable scene queries aggregate these features into a compact scene latent, supervised by auxiliary Gaussian reconstruction and reused across camera-time queries.
A scene-guided pixel diffusion transformer (DiT) cross-attends to this latent and generates RGB images via rectified flow, conditioned on the target camera pose and timestamp.
\method achieves state-of-the-art performance on static indoor (RE10K) and dynamic outdoor (Waymo) benchmarks, with substantial gains in dynamic regions. With scaled training data, \method further demonstrates effective zero-shot generalization to general dynamic scenes, as evaluated on DyCheck and TUM datasets.
\end{abstract}


\section{Introduction}
\label{sec:intro}

Novel view synthesis (NVS) lets us replay a recorded event from a new camera pose, supporting applications including free-viewpoint virtual tourism~\cite{tung2024megascenes,xie2026lavr} and driving simulation~\cite{yang2025storm,zhou2026hugsim,tian2026simscale, zeng2025industrial}.
In dynamic scenes, high-fidelity NVS further requires integrating geometry, appearance, and motion across views and time~\cite{huang2026spacetimepilot, liang2024btimer}.
Many feed-forward methods organize scene evidence around input observations, either as pixel-aligned Gaussian primitives~\cite{charatan2024pixelsplat,liang2024btimer,lin2026movies,yu2026streetforward} or as frame-indexed latent features for neural rendering~\cite{szymanowicz2026lagernvs,hong2026fl4n,liang2024btimer}.
These observation-aligned representations preserve local image evidence. However, consolidating such evidence across inputs remains challenging: overlapping views may encode redundant scene content, whereas moving objects exhibit evolving states across time.

To consolidate redundant evidence across views, scene-aligned approaches use learned queries to aggregate observations into bounded, compact scene representations~\cite{ren2026tokengs,an2025c3g} whose output token count is independent of the source-frame count and resolution. 
Such representations have been used to decode time-specific Gaussian reconstruction~\cite{c4g} and to condition the generative diffusion model~\cite{asim2026scenetok}.
However, preserving effective dynamic content in compact representation and faithfully decoding it into detailed images remain a challenge.

We present \method, a generative dynamic NVS framework built around a compact dynamics-aware scene latent. 
First, the scene encoder uses Temporal-Projective RoPE (TP-RoPE) to incorporate camera and temporal context into complementary geometry and appearance features. Then learnable scene queries aggregate these features into a fixed-budget scene latent and an auxiliary Gaussian reconstruction supervision provides a straightforward training signal for learning the compact scene latent.
Finally, a scene-guided pixel DiT directly generates RGB images. It accesses the scene latent through cross-attention, reusing the representation across target queries.

Our contributions are summarized as follows:

\begin{itemize}
    \item We propose \method, a generative NVS framework that leverages a compact, dynamics-aware scene latent to jointly capture scene geometry and appearance across viewpoints and time.
    \item We develop a scene-guided pixel DiT decoder that generates RGB images directly from noise, supporting independent control over both camera pose and requested timestamp.
    \item We demonstrate substantial improvements in NVS quality over prior methods across static and dynamic scenes, while retaining a compact, fixed-budget scene representation.
\end{itemize}

\section{Related Work}
\label{sec:rel_work}

\paragraph{Dynamic Scene Reconstruction}
Per-scene methods reconstruct motion by fitting deformable radiance fields or time-varying Gaussians to each sequence~\cite{pumarola2021dnerf,park2021nerfies,yang2024deformable,4dgs,yang2024gs4d,li2024spacetime}. Feed-forward models predict dynamic Gaussians directly from video~\cite{lin2025dgslrm,xu20254dgt,ma20254dlrm}, including in street scenes with moving traffic~\cite{yang2025storm,chen2026dggt}. StreetForward~\cite{yu2026streetforward} uses predicted velocities to aggregate evidence across frames. 4RC~\cite{luo20264rc} supports 4D reconstruction through conditional space--time querying. The concurrent work, FL4N~\cite{hong2026fl4n}, learns a temporal residual over a frozen static NVS latent and adapts it to the target timestamp before rendering at the requested camera pose. Despite cross-frame attention, its static and residual latents retain a patch grid per source frame. We instead aggregate observations into a fixed-budget scene latent, rather than retaining frame-indexed conditioning tokens.

\paragraph{Compact 3D/4D Scene Representations}
Compact representations encode scene information beyond a one-to-one correspondence with input pixels. LeanGaussian~\cite{wu2025leangaussian} and TokenGS~\cite{ren2026tokengs} decouple Gaussian prediction from pixel or point-cloud correspondence, while GlobalSplat~\cite{itkin2026globalsplat} uses global scene tokens. SceneTok~\cite{asim2026scenetok} compresses posed views into unstructured scene tokens. C3G~\cite{an2025c3g} learns compact Gaussian representations for static scenes. C4G~\cite{c4g} conditions a compact set of learnable Gaussian queries on the target timestamp to aggregate multi-frame evidence and model global motion, with a video diffusion model refining splatted renderings. NOVA3R~\cite{chen2026novar} explores non-pixel-aligned tokens for amodal reconstruction. In contrast to C4G's target-time Gaussian decoding, our tokenizer produces a shared 4D scene latent that remains unchanged across target camera--time queries.

\paragraph{Generative Novel-View Synthesis}
Generative novel-view synthesis combines observations, pose conditioning, and learned visual priors to synthesize requested images. Pose-conditioned models illustrate different conditioning mechanisms: GEN3C~\cite{ren2025gen3c} uses a 3D cache, while Rays as Pixels~\cite{jang2026rays} models a joint distribution of videos and camera trajectories. Geometry-aware latent approaches such as 3DRAE~\cite{wei20263drae} and GAE~\cite{lu2026gae}, together with the 4D-aware VideoTok4D tokenizer~\cite{chen2026videotok4d}, further explore scene structure in generative modeling. SceneTok~\cite{asim2026scenetok} couples compact scene tokens with pose-conditioned diffusion rendering in a video-autoencoder latent space. 
Our method instead learns scene tokens from complementary geometry and appearance features with explicit temporal conditioning and auxiliary Gaussian supervision, and renders them directly in RGB space at queried camera poses and timestamps.

\section{Method}
\label{sec:method}

\begin{figure*}[t]
    \centering
    \includegraphics[width=\linewidth]{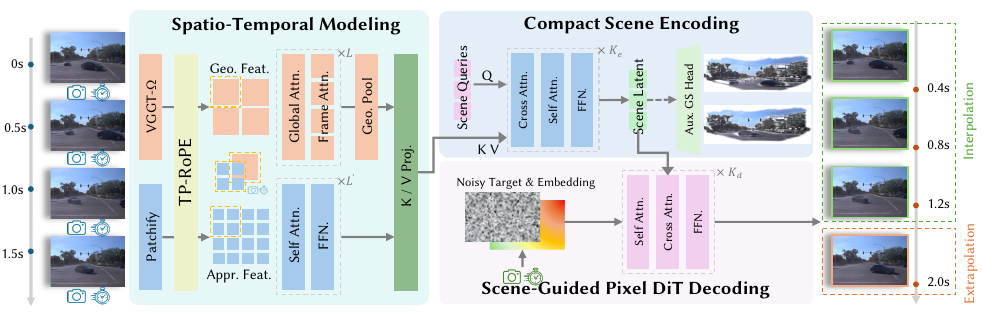}
    \caption{\textbf{Overview of \method.} Geometry and appearance features are conditioned on spatial, temporal, and camera context through TP-RoPE, then aggregated by learnable scene queries into a compact scene representation. An auxiliary Gaussian head provides reconstruction supervision during training. The scene-guided pixel DiT reads the scene tokens through cross-attention and generates RGB images from Gaussian noise, conditioned on the target camera pose and timestamp. The scene representation is computed once and reused across target queries.}
    \label{fig:overview}
\end{figure*}

\method synthesizes novel views of dynamic scenes at specified camera poses and timestamps from sparse posed observations.
As shown in Figure~\ref{fig:overview}, our method separates scene representation from image synthesis.
We first extract geometry and appearance features and incorporate their spatial, temporal, and camera context through TP-RoPE (Section~\ref{sec:spatiotemporal_modeling}).
Learnable scene queries aggregate these features into a compact set of scene tokens, supervised with an auxiliary Gaussian rendering head (Section~\ref{sec:scene_tokenizer}).
A scene-guided pixel DiT decoder then generates target RGB images directly in pixel space, conditioned on the shared scene tokens and the requested camera-time query (Section~\ref{sec:pixel_dit}).
Section~\ref{sec:training} summarizes the scene and pixel supervision objectives.

\subsection{Spatio-Temporal Modeling}
\label{sec:spatiotemporal_modeling}

Given $V$ observations $\mathcal{O}=\{(\mathbf{I}_v,\mathbf{C}_v,t_v)\}_{v=1}^{V}$, where $\mathbf{I}_v$, $\mathbf{C}_v$, and $t_v$ denote an RGB image, its camera parameters, and its physical timestamp, respectively, our model aims to extract both geometric structure and fine-grained appearance information from the observed dynamic scene. 

\paragraph{Geometric and Appearance Feature Extraction.}
We extract geometry-aware features with VGGT-$\Omega$~\cite{wang2026vggtomega}, whose intra-frame and inter-frame attention capture image structure and relationships between observations. Features from layers $\mathcal{L}=\{4,11,17,23\}$ are fused into a geometry stream $\mathbf{F}_{\mathrm{geo}}$, retaining cues from multiple network depths for scene abstraction.

A lightweight ViT-style encoder~\cite{dosovitskiy2021vit} complements the geometry branch with fine-grained appearance features. It uses smaller RGB patches than the geometry branch, providing denser sampling of textures and object boundaries. Self-attention yields the appearance feature stream $\mathbf{F}_{\mathrm{app}}$. 
We then integrate these complementary cues using time-aware embeddings to better capture scene dynamics.

\paragraph{Temporal-Projective RoPE.}
Aggregating features across observations requires preserving their distinct image locations, camera geometries, and capture times. To this end, we extend PRoPE~\cite{li2025prope} with a temporal rotary component, yielding TP-RoPE. For token $i$ from observation $v$ in branch $b\in\{\mathrm{geo},\mathrm{app}\}$, let $(h_i^{(b)},w_i^{(b)})$ denote its position on the branch-specific patch grid, and let $t_v$ and $\mathbf{C}_v$ denote the capture time and camera parameters of observation $v$, respectively. We define the combined transformation as:
\begin{equation}
    \mathbf{D}_{v,i}^{(b)}
    =\mathbf{R}_{h}(h_i^{(b)})\oplus\mathbf{R}_{w}(w_i^{(b)})\oplus\mathbf{R}_{t}(t_v)\oplus\mathbf{P}(\mathbf{C}_v),
\label{eq:tokenizer_tprope}
\end{equation}
where $\mathbf{R}_{h}$ and $\mathbf{R}_{w}$ are spatial rotary operators, and $\mathbf{R}_{t}$ is a temporal rotary operator. $\mathbf{P}(\mathbf{C}_v)$ is PRoPE's projective camera transformation. The direct sum $\oplus$ assigns these components to separate channel subspaces. 
We incorporate TP-RoPE into the attention transformations of both branches, allowing subsequent feature aggregation to account for the spatial, geometric, and temporal context of each observation.

\subsection{Compact Scene Encoding}
\label{sec:scene_tokenizer}



The previous geometry and appearance cues provide complementary information across observations, but their token count grows with the number of input views. We aggregate and compress these cues into a fixed-size scene representation using learnable scene queries. Through cross-attention, the queries gather evidence across views and times, producing shared scene tokens $\mathbf{Z}_{\mathrm{s}}\in\mathbb{R}^{N_{\mathrm{s}}\times d}$ that can be reused for different target camera-time queries.

\paragraph{Scene Token Aggregation.}
We project both feature streams to a common dimension $d$ and concatenate their tokens across all observations:
\begin{equation}
    \mathbf{F}_{\mathrm{obs}}
    =\operatorname{Concat}
    (\widetilde{\mathbf{F}}_{\mathrm{geo}},
     \widetilde{\mathbf{F}}_{\mathrm{app}}),
\end{equation}
where $\widetilde{\mathbf{F}}_b\in\mathbb{R}^{N_b\times d}$ contains the projected tokens from branch $b\in\{\mathrm{geo},\mathrm{app}\}$, with $N_b$ tokens across all observations. A fixed set of learnable queries $\mathbf{Q}^{0}\in\mathbb{R}^{N_{\mathrm{s}}\times d}$ reads from this observation memory over $K_e$ aggregation blocks:
\begin{equation}
    \mathbf{Q}^{\ell+1}
    =\mathcal{A}_{\ell}(\mathbf{Q}^{\ell},\mathbf{F}_{\mathrm{obs}}),
    \ \ \ell=0,\ldots,K_e-1.
\label{eq:scene_query_aggregation}
\end{equation}
Each block uses cross-attention to progressively integrate evidence across views and time into the fixed query set, followed by self-attention to exchange information among the queries and a feed-forward network. The final representation is $\mathbf{Z}_{\mathrm{s}}=\mathbf{Q}^{K_e}\in\mathbb{R}^{N_{\mathrm{s}}\times d}$, where the number of scene tokens $N_{\mathrm{s}}$ is fixed and independent of the number of input observations. It encodes their time-dependent content in shared tokens, without timestamp-specific latent slots or an explicit motion field, and can be reused for different target camera-time queries.

\paragraph{Auxiliary Gaussian Supervision.}
To encourage the compressed tokens to preserve the scene's geometry and appearance, we apply auxiliary supervision through differentiable Gaussian rendering. For each supervised view, a lightweight feed-forward head takes the scene latent encoded from context views and the view's timestamp to predict time-specific 3D Gaussians, which are rendered at the corresponding camera pose.
Supervision covers both context and target views in the training sequence.
The auxiliary head is used only for scene latent supervision during training.

\subsection{Scene-Guided Pixel DiT Decoding}
\label{sec:pixel_dit}

Although the auxiliary Gaussian head can render the scene at novel views, synthesizing high-quality images remains challenging, particularly under sparse observations and such a compact latent. We therefore use a conditional pixel DiT for image synthesis, separating the supervision of the scene representation from the capacity of the final renderer.
The pixel DiT decoder~\cite{peebles2023dit} contains $K_d$ layers of Transformer blocks, which generate RGB images conditioned on the scene tokens $\mathbf{Z}_{\mathrm{s}}$ and a target camera-time query $(\mathbf{C}^{\star},t^{\star})$.

\paragraph{Pixel-Space Generation.}
Following JiT~\cite{li2025jit}, we use clean-image prediction directly in RGB space, avoiding an additional bottleneck introduced by latent space decoding. For a target image $\mathbf{x}_1\in\mathbb{R}^{3\times H\times W}$ normalized to $[-1,1]$:
\begin{equation}
    \mathbf{x}_u=(1-u)\boldsymbol{\epsilon}+u\mathbf{x}_1, \ \
    \boldsymbol{\epsilon}\sim\mathcal{N}(\mathbf{0},\mathbf{I}),
\label{eq:pixel_flow_path}
\end{equation}
where flow progress $u\in[0,1]$ is distinct from physical scene time $t^{\star}$ and modulates the DiT blocks through Adaptive Layer Normalization (AdaLN). The clean-image prediction $\widehat{\mathbf{x}}_1$ gives the stabilized velocity:
\begin{equation}
    \mathbf{v}_{\theta}(\mathbf{x}_u,u)
    =\frac{\widehat{\mathbf{x}}_1-\mathbf{x}_u}{D(u)},
    \ \ D(u)\triangleq\max\{1-u,\eta\},
\label{eq:pixel_flow_velocity}
\end{equation}
with $\eta>0$ stabilizing the denominator near $u=1$. Sampling integrates this velocity from Gaussian RGB noise, without a separate latent-to-RGB decoder.

\paragraph{Camera-Time Query Conditioning.}
We encode the target camera with Pl\"ucker ray-patch embeddings~\cite{zhang2024gslrm} and the relative target time $\Delta t^{\star}=t^{\star}-\max_v t_v$ (in seconds) with a learned embedding shared across patches, adding both to the noisy-image patch tokens.

\paragraph{Scene-Token Conditioning.}
Each DiT block applies image self-attention, scene cross-attention, and a feed-forward network. Cross-attention uses the camera-time-conditioned image features $\mathbf{H}^{\ell}$ as queries and a learned token-wise projection of $\mathbf{Z}_{\mathrm{s}}$, denoted $\mathbf{M}_{\mathrm{s}}$, as KV cache:
\begin{equation}
\mathbf{Q}_{\mathrm{img}}^{\ell}=\mathbf{H}^{\ell}\mathbf{W}_{Q}^{\ell},\;
\mathbf{K}_{\mathrm{s}}^{\ell}=\mathbf{M}_{\mathrm{s}}\mathbf{W}_{K}^{\ell},\;
\mathbf{V}_{\mathrm{s}}^{\ell}=\mathbf{M}_{\mathrm{s}}\mathbf{W}_{V}^{\ell}
\label{eq:scene_cross_attention}
\end{equation}
where $\mathbf{W}_{Q}^{\ell}$, $\mathbf{W}_{K}^{\ell}$, and $\mathbf{W}_{V}^{\ell}$ are learned query, key, and value projections at layer $\ell$. Only image tokens are updated; scene memory is fixed across layers and sampling steps and reused across target queries. The output head predicts RGB patches that form $\widehat{\mathbf{x}}_1$.

\subsection{Training Objectives}
\label{sec:training}

We train the scene representation and pixel decoder with complementary scene and pixel losses.

\paragraph{Scene Loss.}
The scene loss supervises auxiliary Gaussian reconstruction using images from the full sequence, including both context and target views, encouraging the scene tokens to retain appearance and structure:
\begin{equation}
\mathcal{L}_{\mathrm{scene}}=\mathcal{L}_{\mathrm{rgb}}+\lambda_{\mathrm{ssim}}\mathcal{L}_{\mathrm{ssim}}+\lambda_{\mathrm{proj}}\mathcal{L}_{\mathrm{proj}}+\lambda_{\mathrm{depth}}\mathcal{L}_{\mathrm{depth}}.
\label{eq:encoder_pretraining_loss}
\end{equation}
The terms are RGB MSE, SSIM~\cite{wang2004ssim}, a projection penalty for Gaussians outside all supervised views, and masked $L_1$ depth error where ground-truth depth is available.

\paragraph{Pixel Loss.} The pixel loss supervises target image synthesis quality using JiT's velocity-space objective~\cite{li2025jit}:
\begin{equation}
    \mathcal{L}_{\mathrm{pixel}}
    = \mathbb{E}\left[
        \frac{\operatorname{MSE}(\widehat{\mathbf{x}}_1,\mathbf{x}_1)}{D(u)^2}
    \right].
    \label{eq:pixel_flow_training_loss}
\end{equation}
Here $D(u)$ is defined in Equation~\ref{eq:pixel_flow_velocity}. When depth annotations are available, $\mathbf{x}_1$ includes a depth channel; MSE weights all channels equally and masks invalid depth pixels.
The total loss is: 
\begin{equation}
    \mathcal{L}_{\mathrm{total}}=\mathcal{L}_{\mathrm{scene}}+\mathcal{L}_{\mathrm{pixel}}.
\label{eq:total_loss}
\end{equation}

\section{Experiments}
\label{sec:experiments}

\begin{figure*}
    \centering
    \includegraphics[width=\textwidth]{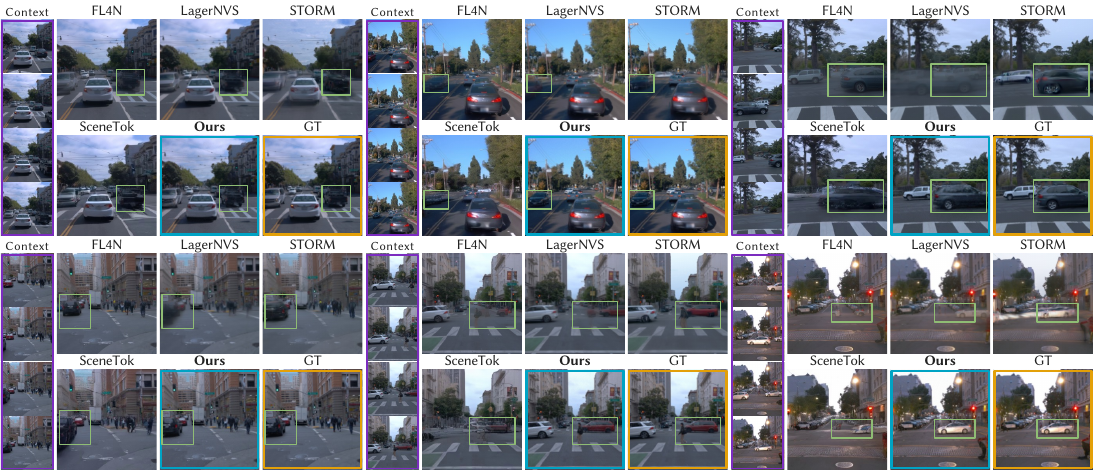}
    \caption{\textbf{Qualitative NVS comparisons on Waymo.}  Results are shown for the 4-context, 16-target setting, with one target view displayed per sequence. (\colorboxmark{contextpurple}, \colorboxmark{predcyan}, \colorboxmark{gtyellow}) denote the context views, our predictions, and ground-truth target views, respectively. Green boxes highlight moving objects. Compared with the baselines, \method produces sharper renderings and better preserves object structure in dynamic regions using a compact scene representation.}
    \label{fig:waymo_results}
\end{figure*}

\subsection{Experimental Setup}
\label{subsec:experimental_setup}

\paragraph{Implementation Details.}
All images are resized to $256\times256$.
We initialize the geometry encoder from pretrained VGGT-$\Omega$~\cite{wang2026vggtomega}.
The default scene latent contains $1{,}024$ tokens of dimension $1{,}024$.
We train with AdamW and a one-cycle learning-rate schedule.
Inference uses 50 steps with a final Euler update~\cite{li2025jit}.
Detailed architecture, conditioning, and training settings are provided in the \textit{Supplementary Materials}.

\paragraph{Datasets.}
Our training datasets include Waymo~\cite{waymo}, RealEstate10K (RE10K)~\cite{re10k}, SpatialVid~\cite{wang2025spatialvid} and nuScenes~\cite{caesar2020nuscenesmultimodaldatasetautonomous}.
For the Waymo and RE10K benchmarks, we train separate models on their respective datasets.
For zero-shot dynamic NVS evaluation on DyCheck~\cite{gao2022dynamic} and TUM~\cite{sturm12iros} datasets, we train the model jointly on RE10K and SpatialVid.
The data-scaling variant combines all four datasets.
For Waymo and SpatialVid, we sample 4 context views at a temporal stride of 4 from each 20-frame clip and evaluate NVS on the remaining 16 frames, covering both temporal interpolation and extrapolation.
RE10K follows SceneTok's 12-context, 128-target and 5-context, 64-target protocols~\cite{asim2026scenetok}.
Dataset splits, frame sampling, and preprocessing are detailed in the \textit{Supplementary Materials}.

\paragraph{Baseline Methods.}
We categorize baselines by representation. Explicit methods render predicted geometric Gaussian primitives, which include DA3~\cite{da3}, DGGT~\cite{chen2026dggt}, StreetForward~\cite{yu2026streetforward}, C3G~\cite{an2025c3g}, and C4G~\cite{c4g}. Implicit methods synthesize images from learned features or tokens, which include LVSM~\cite{jin2025lvsm}, LagerNVS~\cite{szymanowicz2026lagernvs}, SceneTok~\cite{asim2026scenetok}, and concurrent work, FL4N~\cite{hong2026fl4n}. Our method is also implicit. Dataset-specific baselines are listed in Tables~\ref{tab:driving_nvs}, \ref{tab:re10k}, and~\ref{tab:dynamic_nvs}. An asterisk marks our implementation. 
\textit{Repr. Size} counts the 32-bit floating-point values retained per scene: primitive attributes and auxiliary state for explicit methods, or context tokens for implicit methods.

\subsection{Novel-view Synthesis Evaluation}

\begin{figure*}
    \centering
    \includegraphics[width=\linewidth]{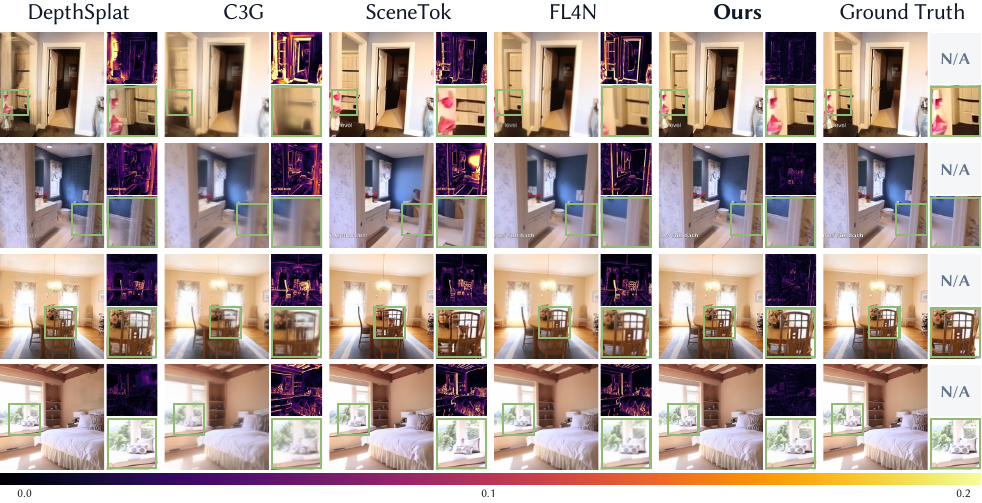}
    \caption{\textbf{Qualitative NVS comparisons on RE10K.} Results are shown for the 12-context, 128-target setting, with one target view displayed per scene. We present the error heatmap between each prediction and the ground truth (top right) as well as the zoomed-in view of local details (bottom right).}
    \label{fig:re10k}
\end{figure*}

\paragraph{NVS on Autonomous Driving Scenes.}
\label{par:driving_nvs}

\begin{table}[t]
    \centering
    \caption{\textbf{Full-image and dynamic-region RGB synthesis on Waymo.} Our methods are shaded gray; best and second-best scores are bold and underlined, respectively. * denotes our baseline implementation. $\dagger$ denotes training with additional datasets.}
    \label{tab:driving_nvs}
    \begingroup
    \footnotesize
    \setlength{\tabcolsep}{2pt}
    \renewcommand{\arraystretch}{1.12}
    \begin{tabularx}{\linewidth}{>{\raggedright\arraybackslash}p{0.35\linewidth}c*{4}{>{\centering\arraybackslash}X}}
        \toprule
        \multirow{2}{*}{Method}
        & \multirow{2}{*}{\shortstack{Repr. Size\\(\#Floats)}}
        & \multicolumn{2}{c}{Full Image}
        & \multicolumn{2}{c}{Dynamic Only} \\
        \cmidrule(lr){3-4}\cmidrule(lr){5-6}
        & & PSNR$\uparrow$ & SSIM$\uparrow$
          & PSNR$\uparrow$ & SSIM$\uparrow$ \\
        \midrule
        \multicolumn{6}{@{}l}{\textit{Explicit Representation}} \\
        AnySplat~\cite{jiang2025anysplat} & 10.61M & 23.32 & 0.687 & 15.99 & 0.418 \\
        DA3 w/ GS Head~\cite{da3} & 15.93M & 23.27 & 0.686 & 15.96 & 0.429 \\
        STORM~\cite{yang2025storm} & 3.15M & 26.38 & 0.794 & 22.10 & 0.624 \\
        DGGT~\cite{chen2026dggt} & 15.93M & 27.41 & 0.846 & 20.99 & 0.821 \\
        StreetForward~\cite{yu2026streetforward} & 2.65M & 27.01 & 0.818 & 24.30 & 0.827 \\
        C3G~\cite{an2025c3g} & 4.19M & 15.81 & 0.577 & 15.48 & 0.224 \\
        C4G~\cite{c4g} & 4.19M & 16.10 & 0.606 & 15.61 & 0.142 \\
        \midrule
        \multicolumn{6}{@{}l}{\textit{Implicit Representation}} \\
        LagerNVS~\cite{szymanowicz2026lagernvs} & 4.21M & 24.12 & 0.720 & 19.82 & 0.390 \\
        SceneTok~\cite{asim2026scenetok} & 0.06M & 26.51 & 0.795 & 21.37 & 0.532 \\
        SceneTok* (Scaled) & 1.05M & 28.27 & 0.833 & 22.80 & 0.606 \\
        FL4N~\cite{hong2026fl4n} & 8.41M & 25.48 & 0.763 & 21.02 & 0.463 \\
        \midrule
        \rowcolor{gray!12}
        \method -Tiny & 0.06M & 26.86 & 0.762 & 21.85 & 0.541 \\
        \rowcolor{gray!12}
        \method -Small & 0.26M & 30.20 & 0.849 & 25.33 & 0.728 \\
        \rowcolor{gray!12}
        \textbf{\method} & 1.05M & \underline{32.91} & \underline{0.888} & \underline{28.77} & \underline{0.838} \\
        \rowcolor{gray!12}
        \method $^{\dagger}$ & 1.05M & \textbf{33.55} & \textbf{0.901} & \textbf{29.45} & \textbf{0.861} \\
        \bottomrule
    \end{tabularx}
    \endgroup
\end{table}

We first evaluate the dynamic NVS on Waymo dataset. Table~\ref{tab:driving_nvs} shows that \method outperforms the compared explicit and implicit baselines in both full-image and dynamic-region PSNR and SSIM, with a smaller per-scene representation than the explicit methods. The improvements in dynamic regions highlight the effectiveness of our compact scene representation in modeling moving objects. Notably, even after scaling SceneTok's scene representation to match ours at 1.05M floats, \method still substantially outperforms it across both full-image and dynamic-region metrics. As shown in Figure~\ref{fig:waymo_results}, our method better preserves the shapes and boundaries of moving vehicles and pedestrians, with less blurring and ghosting in the highlighted regions.

\paragraph{NVS on Static Indoor Scene.}
\label{par:static_nvs}
We evaluate static NVS on RE10K using the same evaluation settings as SceneTok~\cite{asim2026scenetok}. As shown in Table~\ref{tab:re10k}, \method achieves competitive reconstruction quality at the same compact representation size of 0.03M floats. Increasing the representation to 1.05M floats brings substantial gains in NVS quality for \method, whereas SceneTok benefits only modestly from the additional capacity. These results support the effectiveness and scalability of our scene representation. Figure~\ref{fig:re10k} further shows that \method preserves fine structural details, while maintaining the overall appearance of the scene.

\begin{table*}[t]
    \centering
    \caption{\textbf{Novel-view synthesis on RE10K.} Baseline scores follow the results reported in SceneTok~\cite{asim2026scenetok}.}
    \label{tab:re10k}
    \begingroup
    \footnotesize
    \renewcommand{\arraystretch}{1.12}
    \begin{tabularx}{\linewidth}{@{}>{\raggedright\arraybackslash}p{0.18\linewidth}*{12}{@{}>{\centering\arraybackslash}X}@{}}
        \toprule
        & \multicolumn{6}{c}{12 Context Views $\rightarrow$ 128 Target Views}
        & \multicolumn{6}{c}{5 Context Views $\rightarrow$ 64 Target Views} \\
        \cmidrule(lr){2-7}\cmidrule(lr){8-13}
        Method
        & Repr. Size
        & PSNR$\uparrow$
        & SSIM$\uparrow$
        & LPIPS$\downarrow$
        & rFVD$\downarrow$
        & rFID$\downarrow$
        & Repr. Size
        & PSNR$\uparrow$
        & SSIM$\uparrow$
        & LPIPS$\downarrow$
        & rFVD$\downarrow$
        & rFID$\downarrow$ \\
        \midrule
        \multicolumn{13}{@{}l}{\textit{Explicit Representation}} \\
        MVSplat~\cite{chen2024mvsplat}
        & 46.40M & 20.13 & 0.755 & 0.255 & 354.46 & 29.89
        & 19.33M & 22.82 & 0.856 & 0.152 & 144.91 & 15.85 \\
        MVSplat360
        & 74.71M & 21.09 & 0.769 & 0.247 & 294.82 & 25.64
        & 31.12M & 23.88 & 0.828 & 0.174 & 136.52 & 18.15 \\
        DepthSplat~\cite{xu2025depthsplat}
        & 46.40M & 21.55 & \underline{0.809} & 0.202 & 204.52 & 21.35
        & 19.33M & 24.45 & \underline{0.884} & \textbf{0.124} & 109.74 & 12.89 \\
        C3G~\cite{an2025c3g}
        & 4.19M & 20.22 & 0.645 & 0.284 & 325.51 & 33.72
        & 4.19M & 20.02 & 0.605 & 0.292 & 306.62 & 33.36 \\
        \midrule
        \multicolumn{13}{@{}l}{\textit{Implicit Representation}} \\
        LVSM~\cite{jin2025lvsm}
        & 1.57M & 21.25 & 0.686 & 0.262 & 211.66 & 26.40
        & 1.57M & 25.74 & 0.831 & 0.140 & 111.14 & 13.37 \\
        SceneTok~\cite{asim2026scenetok}
        & 0.03M & 23.99 & 0.783 & \underline{0.159} & 79.80 & 11.12
        & 0.03M & 25.97 & 0.817 & \underline{0.133} & 76.24 & 11.26 \\
        SceneTok* (Scaled)
        & 1.05M & 24.08 & 0.790 & \textbf{0.156} & \underline{73.85} & \textbf{9.45}
        & 1.05M & 25.31 & 0.794 & 0.151 & 80.05 & \underline{10.45} \\
        \midrule
        \rowcolor{gray!12}[0pt][0pt]
        \textbf{\method-Tiny}
        & 0.03M
        & \underline{25.38}
        & 0.789
        & 0.243
        & 95.95
        & 16.97
        & 0.03M
        & \underline{26.50}
        & 0.787
        & 0.227
        & \underline{75.85}
        & 19.22 \\
        \rowcolor{gray!12}[0pt][0pt]
        \textbf{\method}
        & 1.05M
        & \textbf{28.03}
        & \textbf{0.845}
        & 0.189
        & \textbf{48.15}
        & \underline{10.53}
        & 1.05M
        & \textbf{28.68}
        & \textbf{0.893}
        & 0.177
        & \textbf{38.83}
        & \textbf{10.08} \\
        \bottomrule
    \end{tabularx}%
    \endgroup
\end{table*}

\paragraph{Zero-shot NVS Evaluation on General Dynamic Scenes.}

We evaluate the zero-shot generalization of \method, trained on SpatialVid~\cite{wang2025spatialvid} and RE10K~\cite{re10k}, to real-world dynamic scenes from DyCheck~\cite{gao2022dynamic} and TUM~\cite{sturm12iros}, without dataset-specific finetuning or per-scene optimization.
Following prior evaluations~\cite{hong2026fl4n}, we use 5 DyCheck scenes and 9 TUM scenes. For each scene, we select the first valid 32-frame clip sampled at 5 FPS, then center-crop and resize the frames to $256\times256$.
All methods use the same clips, with 8 context frames sampled at a stride of 4.
Under this shared protocol, \method consistently outperforms the compared baselines on TUM and achieves competitive overall quality on DyCheck, as shown in Table~\ref{tab:dynamic_nvs}.
These results demonstrate the transferability of our compact, dynamics-aware scene representation and its effectiveness for novel view synthesis in unseen dynamic scenes.

\begin{table}[t]
    \centering
    \caption{\textbf{Zero-shot evaluation on dynamic novel-view synthesis dataset.}}
    \label{tab:dynamic_nvs}
    \begingroup
    \footnotesize
    \setlength{\tabcolsep}{1pt}
    \renewcommand{\arraystretch}{1.12}
    \begin{tabularx}{\linewidth}{>{\raggedright\arraybackslash}p{0.25\linewidth}*{6}{>{\centering\arraybackslash}X}}
        \toprule
        Method & \multicolumn{3}{c}{DyCheck} & \multicolumn{3}{c}{TUM} \\
        \cmidrule(lr){2-4}\cmidrule(lr){5-7}
        & \mbox{PSNR$\uparrow$} & \mbox{SSIM$\uparrow$} & \mbox{LPIPS$\downarrow$}
        & \mbox{PSNR$\uparrow$} & \mbox{SSIM$\uparrow$} & \mbox{LPIPS$\downarrow$} \\
        \midrule
        \multicolumn{7}{@{}l}{\textit{Explicit Representation}} \\
        MoVieS~\cite{lin2026movies}
        & 17.26 & \textbf{0.560} & \underline{0.335} & 17.61 & \underline{0.647} & \underline{0.279} \\
        C4G~\cite{c4g}
        & 13.04 & 0.318 & 0.632
        & 12.46 & 0.288 & 0.599 \\
        \midrule
        \multicolumn{7}{@{}l}{\textit{Implicit Representation}} \\
        LagerNVS~\cite{szymanowicz2026lagernvs}
        & \underline{18.98} & 0.488 & 0.473
        & \underline{18.79} & 0.559 & 0.399 \\
        FL4N~\cite{hong2026fl4n}
        & 18.87 & 0.493 & 0.431
        & 18.68 & 0.552 & 0.383 \\
        \rowcolor{gray!12}
        \textbf{\method}
        & \textbf{19.71} & \underline{0.530} & \textbf{0.304}
        & \textbf{21.79} & \textbf{0.695} & \textbf{0.235} \\
        \bottomrule
    \end{tabularx}
    \endgroup
\end{table}

\subsection{Ablation Studies}
\label{subsec:ablations}

All experiments in this section are trained and tested on the Waymo dataset~\cite{waymo} without special statement.

\begin{figure}[t]
    \centering
    \includegraphics[width=\linewidth]{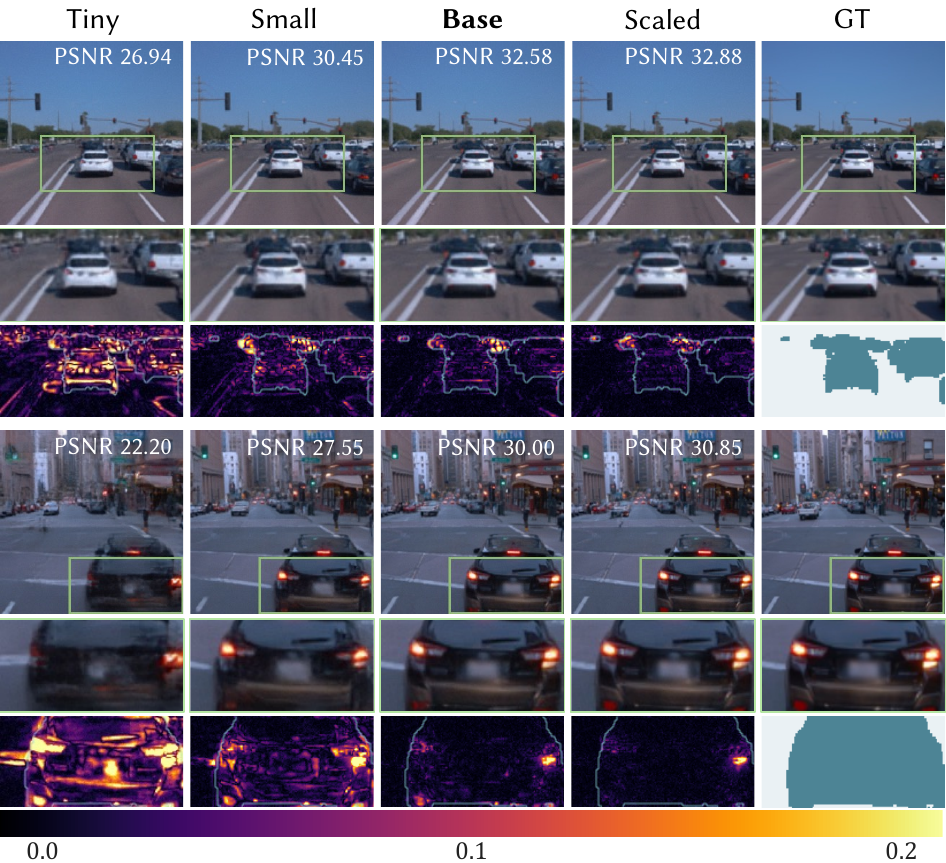}
    \caption{\textbf{Effects of latent compression and data scaling.} Larger scene representations and data-scaled training improve target view fidelity especially for dynamic foreground objects. Rows show full images, enlarged crops, and error heatmaps with a shared scale. GT error-row cells show foreground masks.}
    \label{fig:ablation1}
\end{figure}

\paragraph{Latent Compression and Data Scaling.}
As shown in Table~\ref{tab:driving_nvs}, the smaller \method-Tiny and \method-Small variants explore the trade-off between visual quality and storage cost through scene-latent compression. They outperform baselines with similarly sized representations, indicating more effective use of a compact latent budget.
The final row shows that training on all datasets, including the additional nuScenes dataset, further improves visual quality at a fixed latent size, demonstrating scalability without increasing representation size. 
Figure~\ref{fig:ablation1} provides detailed visual comparisons.

\paragraph{Context Capacity and Temporal Coverage.}
Table~\ref{tab:context_temporal_ablation} examines the effects of varying the number of context views and the temporal span while keeping the scene-latent size fixed. Our representation remains effective under sparse-view inputs, while additional context further improves both full-image and dynamic-region synthesis quality. The same compact latent also supports inference at more target timestamps over longer temporal intervals. Although sparser temporal observations make dynamic content more challenging to reconstruct, the representation remains useful across different temporal coverage settings. These results demonstrate the flexibility of our fixed-budget scene representation in accommodating sparse observations and supporting synthesis across a broader range of target views and times.

\begin{table}[t]
    \centering
    \caption{\textbf{Effect of context capacity and temporal coverage.}
    We evaluate quality with a fixed-size
    scene latent under varying numbers of context frames
    and temporal intervals.}
    \label{tab:context_temporal_ablation}
    \begingroup
    \footnotesize
    \setlength{\tabcolsep}{1pt}
    \renewcommand{\arraystretch}{1.12}
    \begin{tabularx}{\linewidth}{
        >{\centering\arraybackslash}p{0.13\linewidth}
        >{\centering\arraybackslash}p{0.15\linewidth}
        *{6}{>{\centering\arraybackslash}X}
    }
        \toprule
        & &
        \multicolumn{3}{c}{Full Image}
        & \multicolumn{3}{c}{Dynamic Only} \\
        \cmidrule(lr){3-5}\cmidrule(lr){6-8}
        \#Context
        & \#Target
        & \mbox{PSNR$\uparrow$}
        & \mbox{SSIM$\uparrow$}
        & \mbox{LPIPS$\downarrow$}
        & \mbox{PSNR$\uparrow$}
        & \mbox{SSIM$\uparrow$}
        & \mbox{LPIPS$\downarrow$} \\
        \midrule
        \multicolumn{8}{@{}l}{\textit{Baseline}} \\
        \rowcolor{gray!12}
        4 & 16 & \underline{32.91} & \underline{0.888} & \underline{0.229} & \underline{28.77} & \underline{0.838} & \underline{0.162} \\
        \midrule
        \multicolumn{8}{@{}l}{\textit{Context Capacity}} \\
        2 & 18 & 30.48 & 0.862 & 0.256 & 25.91 & 0.755 & 0.214 \\
        8 & 12 & \textbf{33.17} & \textbf{0.894} & \textbf{0.224} & \textbf{28.99} & \textbf{0.849} & \textbf{0.153} \\
        \midrule
        \multicolumn{8}{@{}l}{\textit{Temporal Spacing}} \\
        4 & 24 & 31.99 & 0.879 & 0.239 & 27.66 & 0.812 & 0.181 \\
        4 & 32 & 31.03 & 0.866 & 0.252 & 26.55 & 0.776 & 0.202 \\
        \bottomrule
    \end{tabularx}
    \endgroup
\end{table}

\paragraph{Encoder Architecture.}
Table~\ref{tab:decoder_ablation} replaces TP-RoPE with 2D RoPE and removes either feature branch to assess spatio-temporal conditioning and the complementarity of geometry and appearance features.
The results demonstrate the necessity of our designed embedding mechanism and dual-branch features.
To evaluate the effectiveness of our scene encoder, we further replace it with the encoder from the scaled SceneTok variant in Table~\ref{tab:driving_nvs} and jointly train it with the same scene-guided pixel DiT decoder under identical settings. Our scene encoder yields higher NVS quality under the same decoder architecture, which further supports the effectiveness of scene representation.

\begin{figure}[t]
    \centering
    \includegraphics[width=\linewidth]{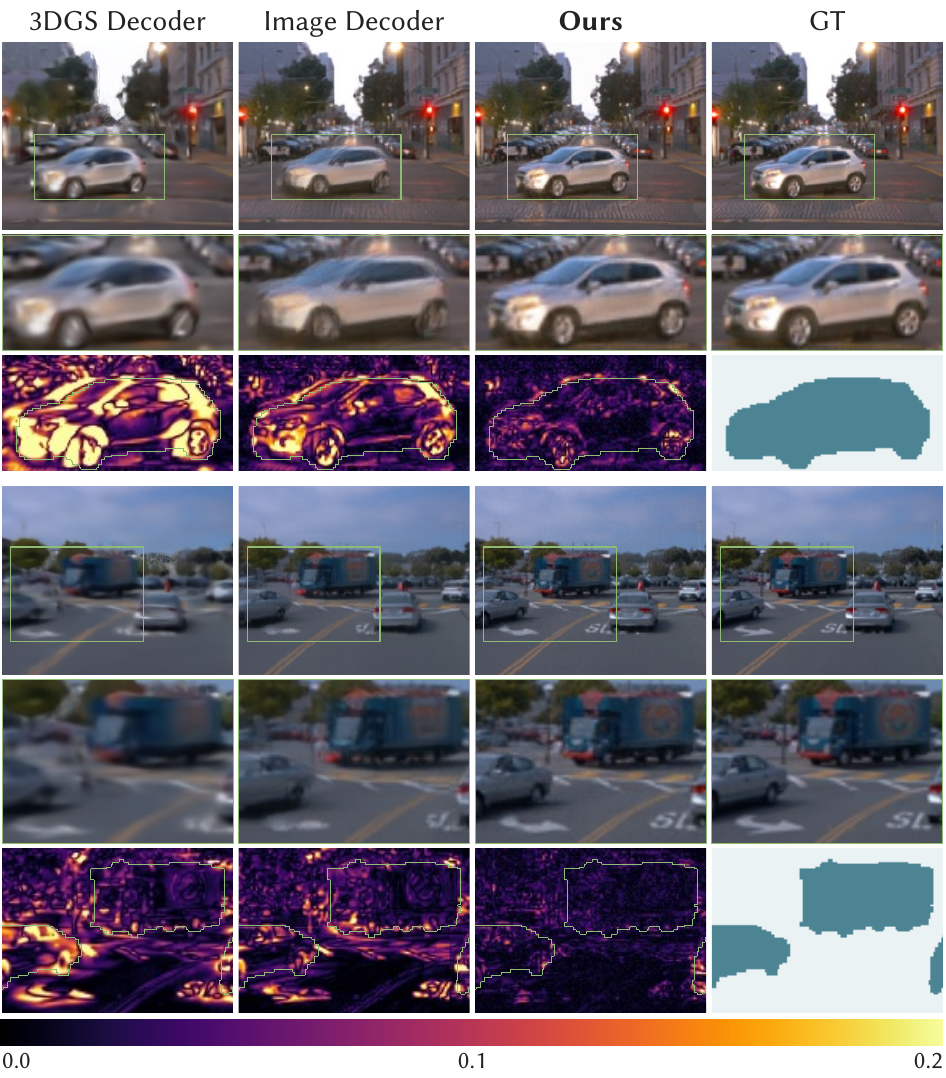}
    \caption{\textbf{Qualitative comparison of decoder architectures.} With the scene encoder jointly fine-tuned, Pixel DiT better preserves dynamic object details than the 3DGS decoder and image decoder. Each example shows full images, enlarged crops, and error heatmaps with a shared scale. GT error-row cells show foreground masks.}
    \label{fig:ablation2}
\end{figure}

\paragraph{Decoder Architecture.}
We replace Pixel DiT with either the auxiliary GS head from Section~\ref{sec:scene_tokenizer} or a LagerNVS-style image decoder~\cite{szymanowicz2026lagernvs} with target-time embeddings, and jointly fine-tune the scene encoder with each decoder. Both recover scene structure and appearance, providing evidence that the compact latent retains both types of information (Table~\ref{tab:decoder_ablation} and Figure~\ref{fig:ablation2}). Our final Pixel DiT achieves higher full-image and dynamic-region PSNR, while the lightweight decoders provide alternatives for balancing inference cost and synthesis quality. Inference efficiency comparisons are provided in the \textit{Supplementary Materials}.

\paragraph{Training Scheme.} Table~\ref{tab:decoder_ablation} separately removes scene reconstruction supervision and the depth channel to assess their effects on synthesis quality on Waymo. Removing $\mathcal{L}_{\mathrm{scene}}$ severely degrades synthesis quality, making it difficult to recover the structure of dynamic scenes. Joint depth denoising provides geometric supervision and improves full-image and dynamic-region synthesis quality.

\paragraph{More Ablations and Discussions.}
Please check our \textit{Supplementary Materials} for further discussion about the component effectiveness, compact representation, and inference efficiency.

\begin{table}[t]
    \centering
    \caption{\textbf{Ablation of model architectures.}}
    \label{tab:decoder_ablation}
    \begingroup
    \footnotesize
    \setlength{\tabcolsep}{1pt}
    \renewcommand{\arraystretch}{1.12}
    \begin{tabularx}{\linewidth}{
        >{\raggedright\arraybackslash}p{0.25\linewidth}
        *{6}{>{\centering\arraybackslash}X}
    }
        \toprule
        & \multicolumn{3}{c}{Full Image}
        & \multicolumn{3}{c}{Dynamic Only} \\
        \cmidrule(lr){2-4}\cmidrule(lr){5-7}
        & \mbox{PSNR$\uparrow$}
        & \mbox{SSIM$\uparrow$}
        & \mbox{LPIPS$\downarrow$}
        & \mbox{PSNR$\uparrow$}
        & \mbox{SSIM$\uparrow$}
        & \mbox{LPIPS$\downarrow$} \\
        \midrule
        \multicolumn{7}{@{}l}{\textit{Scene Modeling Architecture}} \\
        SceneTok Enc.~\cite{asim2026scenetok} & 27.31 & 0.782 & 0.315 & 22.20 & 0.556 & 0.328 \\
        w/o TP-RoPE & 30.99 & 0.826 & 0.289 & 27.08 & 0.785 & 0.393 \\
        w/o Appr. Feat. & 27.77 & 0.704 & 0.405 & 24.54 & 0.529 & 0.418 \\
        w/o Geom. Feat. & 25.15 & 0.691 & 0.532 & 20.73 & 0.596 & 0.573 \\
        \midrule
        \multicolumn{7}{@{}l}{\textit{Decoder Architecture}} \\
        3DGS Decoder
        & 28.92 & 0.860 & \underline{0.222} & 25.31 & \underline{0.829} & 0.245 \\
        Image Decoder
        & 31.23 & \textbf{0.894} & \textbf{0.158} & 26.00 & 0.747 & 0.219 \\
        \midrule
        \multicolumn{7}{@{}l}{\textit{Training Settings}} \\
        w/o $\mathcal{L}_{\textrm{scene}}$ & 19.65 & 0.583 & 0.516 & 16.78 & 0.237 & 0.554 \\
        w/o depth channel
        & \underline{32.15} & 0.870 & 0.257 & \underline{27.86} & 0.812 & \underline{0.204} \\
        \midrule
        \rowcolor{gray!12}
        \textbf{ChronoLatent}
        & \textbf{32.91} & \underline{0.888} & 0.229
        & \textbf{28.77} & \textbf{0.838} & \textbf{0.162} \\
        \bottomrule
    \end{tabularx}
    \endgroup
\end{table}

\section{Conclusion}
\label{sec:conclusion}

We introduced \method for dynamic novel-view synthesis, combining compact spatio-temporal scene encoding with pixel-space generative rendering. By consolidating observations across views and times into scene-level memory, \method supports image synthesis at requested camera poses and timestamps without retaining frame-indexed conditioning features. 
Experiments demonstrate improved generation quality on static and dynamic benchmarks, with substantial gains in dynamic regions and effective zero-shot generalization to general dynamic datasets.


\paragraph{Limitations and Future Work.}
Iterative pixel-space sampling incurs higher inference cost than lightweight decoders. High-fidelity novel-view synthesis in scenes with non-rigid motion also remains challenging. Future work will explore better modeling of non-rigid motion and reducing sampling cost.

\paragraph{LLM Usage Statement.} An LLM-based writing assistant (ChatGPT) was used solely to assist with drafting, editing, and polishing the manuscript based on author-provided content. It was not used to develop the research methodology, conduct experiments, or generate numerical results. All scientific content and findings were provided and verified by the authors, who take full responsibility for the final manuscript. 

{
    \small
    \bibliographystyle{ieeenat_fullname}
    \bibliography{main}
}

\clearpage
\appendix
\section{Implementation Details}
\label{sec:supp_implementation}
We provide detailed model and training configurations for reproducibility.
Section~\ref{sec:supp_architecture} describes the \method architecture, including the feature encoders, compact scene encoding, auxiliary Gaussian head, and scene-guided pixel DiT.
Section~\ref{sec:supp_training_recipe} specifies the training recipe, supervision settings, and inference sampling procedure.

\subsection{ChronoLatent Architecture}
\label{sec:supp_architecture}

Table~\ref{tab:impl_architecture} summarizes the default architecture.
We use a common feature dimension $d=1{,}024$ for scene token aggregation and pixel DiT decoding.
The following details complement the formulation in the main paper.

\begin{table}[t]
    \centering
    \caption{\textbf{Default model configuration.} Images are resized to $H\times W=256\times256$. Attention modules use 16 heads, and FFNs use an expansion ratio of four. Feature-layer indices are zero-based.}
    \label{tab:impl_architecture}
    \begingroup
    \footnotesize
    \setlength{\tabcolsep}{3pt}
    \renewcommand{\arraystretch}{1.08}
    \newcommand{\configgroup}[1]{%
        \rowcolor{gray!10}[0pt][0pt]\multicolumn{2}{@{}l@{}}{\textbf{#1}}\\
    }
    \begin{tabularx}{\linewidth}{@{}>{\raggedright\arraybackslash}p{0.51\linewidth}>{\centering\arraybackslash}X@{}}
        \toprule
        \textbf{Configuration} & \textbf{Value} \\
        \midrule
        \configgroup{Geometry Feature Encoder}
        Hidden width & 1024 \\
        Number of blocks & 24 \\
        Patch size & $16\times16$ \\
        Backbone & VGGT-$\Omega$ \\
        Feature layers & $\{4,11,17,23\}$ \\
        Feature fusion & Mean \\
        Feature projection & $2048\rightarrow1024$ \\
        \midrule
        \configgroup{Appearance Feature Encoder}
        Hidden width & 1024 \\
        Number of blocks & 3 \\
        Patch size & $8\times8$ \\
        \midrule
        \configgroup{Compact Scene Encoding}
        Scene token width $d$ & 1024 \\
        Aggregation blocks $K_e$ & 12 \\
        Scene tokens $N_{\mathrm{s}}$ & 1024 \\
        \midrule
        \configgroup{Auxiliary Gaussian Head}
        Hidden width $d$ & 1024 \\
        Number of blocks & 2 \\
        Block conditioning & AdaLN \\
        Head architecture & Time-conditioned FFN \\
        Output dimension per token & $64\times14$ \\
        Gaussians per timestamp & 65,536 \\
        \midrule
        \configgroup{Scene-Guided Pixel DiT}
        Hidden width $d$ & 1024 \\
        Decoder blocks $K_d$ & 12 \\
        Patch size & $8\times8$ \\
        Block conditioning & AdaLN-Zero \\
        Target camera embedding ($\mathbf{C}^{\star}$) & Pl\"ucker ray patches \\
        Relative-time embedding ($\Delta t^{\star}$) & Fourier features + MLP \\
        Flow-progress embedding ($u$) & Sinusoidal + MLP \\
        RGB output dimension per patch & $8\times8\times3$ \\
        \bottomrule
    \end{tabularx}
    \endgroup
\end{table}

\paragraph{Geometric and Appearance Feature Extraction.}
We initialize the geometry encoder from VGGT-$\Omega$~\cite{wang2026vggtomega} and extract features from layers $\mathcal{L}=\{4,11,17,23\}$.
At each selected layer, concatenating the intra-frame and inter-frame attention outputs gives a 2,048-dimensional patch feature.
We average these features across layers and project them to dimension $d$.
The appearance encoder uses 3-layer ViT blocks with $8\times8$ RGB patches, compared to $16\times16$ patches in the geometry encoder.
At the default resolution, each observation therefore contributes 256 geometry tokens and 1,024 appearance tokens.

\paragraph{Temporal-Projective RoPE.}
TP-RoPE partitions each 64-dimensional attention head into four 16-channel groups corresponding to patch row, patch column, observation time, and camera projection, respectively.
The spatial rotations use integer coordinates on each branch's patch grid, and both spatial and temporal rotations use a frequency base of 100.
All patches from observation $v$ share the same temporal rotation, which is applied only to queries and keys so that their dot products encode time differences between observations.
Following PRoPE~\cite{li2025prope}, the camera-projection component combines the world-to-camera transform with image-normalized intrinsics, with the principal point expressed relative to the image center.

\paragraph{Scene Token Aggregation.}
The projected feature streams are concatenated along the token dimension to form $\mathbf{F}_{\mathrm{obs}}$.
LayerNorm and linear key/value projections prepare these features for cross-attention, with per-head query--key normalization.
The learnable scene queries $\mathbf{Q}^{0}$ are updated by $K_e=12$ blocks of cross-attention, self-attention, and a GELU FFN.
Their final features form $\mathbf{Z}_{\mathrm{s}}\in\mathbb{R}^{N_{\mathrm{s}}\times d}$.
Neither the target camera nor the target timestamp enters this aggregation, allowing the resulting scene latent to be reused across camera-time queries.

\paragraph{Auxiliary Gaussian Supervision.}
The auxiliary head decodes the shared scene latent $\mathbf{Z}_{\mathrm{s}}$ into 3D Gaussians at each supervised timestamp $t$.
For time conditioning, we concatenate the relative time $\Delta t=t-\max_v t_v$ with its multi-frequency sinusoidal encoding.
A two-layer SiLU MLP maps this representation to the time condition $\mathbf{e}_t$.

The head processes each scene token $\mathbf{z}_i$ independently through two time-conditioned residual FFN blocks.
Starting from $\mathbf{h}_i^0=\mathbf{z}_i$, each block updates the token as
\begin{equation}
\mathbf{h}_i^{\ell+1}=\mathbf{h}_i^\ell+
\operatorname{FFN}_{\ell}\!\left(
\operatorname{AdaLN}_{\ell}(\mathbf{h}_i^\ell;\mathbf{e}_t)\right),
\label{eq:supp_gaussian_ffn}
\end{equation}
for $\ell\in\{0,1\}$.
Each FFN is a two-layer MLP with GELU activation.
AdaLN applies non-affine LayerNorm followed by channel-wise scale and shift predicted from $\mathbf{e}_t$ through a block-specific SiLU--linear projection.
The predicted scale is offset by one, and both blocks' modulation projections are zero-initialized.
Head weights are shared across tokens and timestamps.

A final LayerNorm and linear projection produce 64 Gaussians per token, each parameterized by position, scale, quaternion rotation, opacity, and RGB color.
We parameterize positions with a signed exponential, scales with a capped exponential, rotations with normalized quaternions, opacity with a sigmoid, and color with a tanh mapping.
The Gaussians are rendered with \texttt{gsplat}~\cite{ye2025gsplat} at the corresponding supervised camera pose.
Reconstruction losses on context and target views supervise $\mathbf{Z}_{\mathrm{s}}$ through this auxiliary head, which is discarded for pixel DiT inference.

\paragraph{Scene-Guided Pixel DiT Decoding.}
The pixel DiT embeds noisy RGB or RGBD images and six-channel Pl\"ucker ray maps~\cite{sitzmann2021light} using separate $8\times8$ patch projections.
For target-time conditioning, the relative timestamp $\Delta t^{\star}$ is encoded using the same Fourier features as the Gaussian head, followed by an independent two-layer SiLU MLP.
The resulting time embedding and camera-ray embeddings are added to the noisy-image patch tokens.
Flow progress $u$ is encoded through a sinusoidal embedding and a separate two-layer SiLU MLP, and conditions the DiT blocks via AdaLN-Zero.

To incorporate scene information, we linearly project $\mathbf{Z}_{\mathrm{s}}$ into the scene memory $\mathbf{M}_{\mathrm{s}}$.
Each DiT block consists of image self-attention, cross-attention to $\mathbf{M}_{\mathrm{s}}$, and a GELU FFN, with per-head query--key normalization in both attention layers.
Spatial RoPE is applied to image self-attention, while scene cross-attention accesses the shared scene tokens without image-grid positional encoding.
When depth annotations are available, normalized depth is concatenated with RGB, and all four channels are jointly denoised along the same interpolation path.
A final LayerNorm and linear projection predict pixel values for each patch, which are rearranged into $\widehat{\mathbf{x}}_1$.
Its RGB channels form the synthesized image.

\paragraph{Image Decoder for the Decoder Ablation.}
We adopt a LagerNVS-style transformer~\cite{szymanowicz2026lagernvs} with 12 blocks and a hidden dimension of 768.
Patch embeddings of the target-camera Pl\"ucker ray maps~\cite{sitzmann2021light} are combined with register tokens to form the query sequence, which interacts with our scene-token memory through bidirectional cross-attention.
To incorporate target-time conditioning, we encode the relative target timestamp using the Fourier features described in Section~\ref{sec:supp_architecture}, followed by a two-layer SiLU MLP.
The resulting time embedding is projected and added to the query tokens.
Separate block-specific projections map this embedding to channel-wise scale and shift parameters, which modulate the query stream before each block via FiLM~\cite{perez2018film}:
$\mathbf{h}'_{\ell}=\mathbf{h}_{\ell}\odot(1+\boldsymbol{\gamma}_{\ell})+\boldsymbol{\beta}_{\ell}$.
The decoder directly predicts the target RGB image without iterative sampling.
\subsection{Training Recipe}
\label{sec:supp_training_recipe}

\begin{table}[t]
    \centering
    \caption{\textbf{Default training and inference settings on Waymo.}}
    \label{tab:training_configuration}
    \footnotesize
    \setlength{\tabcolsep}{3pt}
    \renewcommand{\arraystretch}{1.08}
    \newcommand{\traininggroup}[1]{%
        \rowcolor{gray!10}[0pt][0pt]\multicolumn{2}{@{}l@{}}{\textbf{#1}}\\
    }
    \begin{tabularx}{\linewidth}{@{}>{\raggedright\arraybackslash}p{0.57\linewidth}>{\centering\arraybackslash}X@{}}
        \toprule
        \textbf{Configuration} & \textbf{Value} \\
        \midrule
        \traininggroup{Optimization}
        Precision & bf16 \\
        Global batch size & 8 \\
        Epochs & 300 \\
        Iterations per epoch & 500 \\
        Optimizer & AdamW \\
        Initial AdamW $(\beta_1,\beta_2)$ & $(0.9,0.95)$ \\
        Weight decay & 0.05 \\
        Gradient clipping & 1.0 \\
        Learning-rate schedule & One-cycle \\
        Warmup steps & 2{,}000 \\
        Scene encoder peak LR & $2\times10^{-4}$ \\
        Pixel DiT peak LR & $2\times10^{-4}$ \\
        VGGT backbone peak LR & $10^{-6}$ \\
        \midrule
        \traininggroup{Flow Matching}
        Prediction parameterization & $\mathrm{x}$-predict \\
        Distribution of flow progress $u$ & Logit-normal \\
        Logit mean & $-0.8$ \\
        Logit standard deviation & $0.8$ \\
        Noise std.\ per component of $\boldsymbol{\epsilon}$ & 1.0 \\
        Weight of $\mathcal{L}_{\mathrm{pixel}}$ & 1.0 \\
        \midrule
        \traininggroup{Supervision}
        SSIM weight $\lambda_{\mathrm{ssim}}$ & 0.2 \\
        Projection weight $\lambda_{\mathrm{proj}}$ & 1.0 \\
        Depth weight $\lambda_{\mathrm{depth}}$ & 0.2 \\
        \midrule
        \traininggroup{Sampling}
        Sampler & Heun with final Euler step \\
        Sampling steps & 50 \\
        Velocity denominator clamp $\eta$ & 0.05 \\
        \bottomrule
    \end{tabularx}
\end{table}

Table~\ref{tab:training_configuration} summarizes the default settings on Waymo.
We train in BF16 on eight NVIDIA H800 GPUs (80\,GB each) using AdamW.
The one-cycle schedule varies the learning rate and $\beta_1$, with a shared peak rate for the scene encoder and pixel DiT and a lower peak rate for the VGGT backbone.
Weight decay excludes one-dimensional tensors and parameters marked for exemption.
We train the pixel DiT with clean-image prediction and apply the depth loss where annotations are available.

At inference, we follow JiT~\cite{li2025jit} and use 50 uniformly spaced steps with Heun updates and a final Euler update.
The velocity denominator is lower-bounded by $\eta=0.05$.

\section{Dataset Details}

\paragraph{Waymo and nuScenes.}
We use front-camera clips from Waymo~\cite{waymo} and nuScenes~\cite{caesar2020nuscenesmultimodaldatasetautonomous}, following the official scene splits for both datasets. We train the Waymo benchmark model on the 798 scenes in the Waymo training split. For the data-scaling experiment, we combine Waymo and nuScenes with RE10K and SpatialVID.

\paragraph{RE10K.}
RealEstate10K~\cite{re10k} provides camera intrinsics and poses for video tours of largely static scenes. The training split contains $66{,}033$ sequences. We select 12 context views per training sequence by farthest-point sampling over camera positions and orientations. We sample eight target frames at random from the full sequence, including the context frames among the candidates.

\paragraph{SpatialVID.}
We train on SpatialVID~\cite{wang2025spatialvid} and RE10K for zero-shot evaluation on two benchmarks for dynamic scenes, TUM~\cite{sturm12iros} and DyCheck~\cite{gao2022dynamic}. Our subset contains $12{,}000$ clips from $9{,}000$ source videos, primarily of indoor scenes and outdoor spaces such as gardens and courtyards. Each source contributes one or two clips, and all clips from a given video belong to the same split.
Each clip consists of RGB frames at $256\times256$ resolution, with associated camera intrinsics, camera poses, and original frame timestamps. Context frames are sampled at regular temporal intervals, while the remaining frames serve as targets for interpolation and extrapolation.

\section{Evaluation Protocols}

\paragraph{Waymo.}
We evaluate $2{,}019$ front-camera clips from 202 Waymo validation scenes. Each clip contains 20 consecutive frames. Following STORM~\cite{yang2025storm} and StreetForward~\cite{yu2026streetforward}, we use four context views per clip and render the remaining target frames at their recorded camera poses and timestamps. Dynamic-region masks are constructed by projecting Waymo's ground-truth 3D bounding boxes of moving objects into each target view. We compute PSNR and SSIM at $256\times256$ resolution and average over all $32{,}304$ target images. For ablations, we additionally report VGG-based LPIPS. 

\paragraph{Waymo temporal consistency.}
For the results in Table~\ref{tab:scene_consistency}, we render all 20 timestamps in each clip from the same four context views, including the context timestamps. We measure warping error over the 19 adjacent frame pairs. For each pair, ground-truth LiDAR scene flow and camera poses map pixels from the later frame to the earlier frame. We compute the mean squared RGB difference between predictions over valid correspondences and average the errors across frame pairs. We also report rFVD by comparing the distributions of rendered and ground-truth clips, using one I3D feature vector per 20-frame clip.

\paragraph{RE10K.}
We follow SceneTok's image preprocessing and released scene and frame indices~\cite{asim2026scenetok}. The 12-context protocol uses 136 test scenes and renders 128 views per scene; the 5-context protocol uses 268 test scenes and renders 64 views per scene. Each rendered sequence includes its context views, leaving 116 and 59 held-out views per scene, respectively.

We encode the context views once per scene and render the full sequence at $256\times256$ resolution. PSNR, SSIM, and VGG-based LPIPS are computed over all rendered views, including the context views, and averaged first within each scene and then across scenes. For rFID, we compare Inception features from all rendered frames with those from the corresponding ground-truth frames. For rFVD, we divide each scene into consecutive 16-frame clips and compare the I3D feature distributions of rendered and ground-truth clips.

\paragraph{TUM and DyCheck.}
Following FL4N~\cite{hong2026fl4n}, we use nine TUM~\cite{sturm12iros} scenes and five DyCheck~\cite{gao2022dynamic} scenes. All methods are evaluated on the same 32-frame clip per scene, sampled at 5~fps from the first valid timestamp; for DyCheck, we use camera~0. We estimate camera poses and intrinsics jointly from all 32 RGB frames with VGGT~\cite{wang2025vggt} and share these estimates across methods.

Each method uses the eight frames at indices $0,4,\ldots,28$ as context and renders all 32 frames at $256\times256$ resolution. We compute full-image PSNR, SSIM, and VGG-based LPIPS over the 29 frames from the first through the last context timestamp (indices 0--28, including the context frames). We average the metrics within each scene and then across scenes for each dataset.

\section{Inference Efficiency and Cost}
We report the inference efficiency and the VRAM cost in Table~\ref{tab:inference_efficiency}. Overall, our method achieves leading visual quality and synthesis accuracy. Its higher end-to-end inference time is largely attributable to the computational cost of the pixel DiT decoder, while its GPU memory footprint remains manageable. However, our image decoder and Gaussian decoder variants offer substantially faster decoding and rendering, respectively, while maintaining leading visual quality, providing practical alternatives with favorable quality-efficiency trade-offs. Further accelerating the pixel DiT decoder through single-step inference or distillation is a promising direction for future work.

\begin{table}[t]
    \centering
    \caption{\textbf{Inference efficiency on Waymo.}
    PSNR is evaluated on full images. Encoding time is measured once per scene. Rendering reports the average time per target frame after scene encoding, excluding encoding time. Total time measures the processing time for each sequence.}
    \label{tab:inference_efficiency}
    \begingroup
    \footnotesize
    \setlength{\tabcolsep}{3pt}
    \renewcommand{\arraystretch}{1.12}
    \resizebox{\linewidth}{!}{%
    \begin{tabular}{lccccc}
        \toprule
        Method & PSNR$\uparrow$ & Enc. (s)$\downarrow$
        & Render (s)$\downarrow$ & Total (s)$\downarrow$
        & VRAM (GB)$\downarrow$ \\
        \midrule
        DepthSplat~\cite{xu2025depthsplat}
        & 23.29 & 0.056 & 0.001 & 0.077 & 1.71 \\
        C3G~\cite{an2025c3g}
        & 15.81 & 0.060 & 0.001 & 0.068 & 4.73 \\
        LagerNVS~\cite{szymanowicz2026lagernvs}
        & 24.12 & 0.094 & 0.008 & 0.217 & 7.18 \\
        SceneTok~\cite{asim2026scenetok}
        & 26.51 & 0.032 & 0.114 & 2.305 & 11.44 \\
        FL4N~\cite{hong2026fl4n}
        & 25.48 & 0.140 & 0.009 & 0.291 & 9.84 \\
        \midrule
        \rowcolor{gray!12}
        \textbf{Ours}
        & 32.91 & 0.060 & 0.425 & 6.860 & 6.69 \\
        \textbf{Ours} - Image Decoder
        & 31.23 & 0.060 & 0.003 & 0.112 & 6.05 \\
        \textbf{Ours} - GS Decoder
        & 28.92 & 0.060 & 0.006 & 0.161 & 8.64 \\
        \bottomrule
    \end{tabular}%
    }
    \endgroup
\end{table}

\section{Scene Consistency Evaluation}
\label{sec:supp_scene_consistency}
We further analyze scene consistency on the Waymo dataset, with a particular focus on dynamic content. As shown in Table~\ref{tab:scene_consistency}, our method achieves the best rFVD score while maintaining a relatively low warping error across the sequence. Although StreetForward~\cite{yu2026streetforward} achieves the strongest dynamic consistency, it relies on bounding-box annotations for dynamic reconstruction, whereas our method requires no additional structural inputs. FL4N~\cite{hong2026fl4n} achieves the lowest full-image warping error but exhibits pronounced temporal jitter and abrupt transitions between chunks throughout the sequence. Please see our \textit{supplementary video} for more quality results.

\begin{table}[t]
    \centering
    \caption{\textbf{Temporal consistency on Waymo.}
    Warping error compares adjacent predictions after alignment with
    ground-truth scene flow, both across the full image and within
    dynamic regions. rFVD compares the distributions of rendered and
    ground-truth clips.}
    \label{tab:scene_consistency}
    \begingroup
    \footnotesize
    \setlength{\tabcolsep}{2pt}
    \renewcommand{\arraystretch}{1.12}
    \begin{tabularx}{\linewidth}{l*{3}{>{\centering\arraybackslash}X}}
        \toprule
        \multirow{2}{*}{Method}
        & \multicolumn{2}{c}{Warp Error ($\times 10^{-4}$)$\downarrow$}
        & \multirow{2}{*}{rFVD$\downarrow$} \\
        \cmidrule(lr){2-3}
        & Full Image & Dynamic Only & \\
        \midrule
        \multicolumn{4}{l}{\textit{Explicit Representation}} \\
        StreetForward~\cite{yu2026streetforward}
        & 5.75 & \textbf{21.5} & 677.71 \\
        \midrule
        \multicolumn{4}{l}{\textit{Implicit Representation}} \\
        SceneTok~\cite{asim2026scenetok}
        & 11.3 & 46.2 & 662.58 \\
        FL4N~\cite{hong2026fl4n}
        & \textbf{3.32} & 24.8 & 108.94 \\
        \midrule
        \rowcolor{gray!12}
        \textbf{\method}
        & \underline{5.69} & \underline{23.7} & \textbf{25.68} \\
        \bottomrule
    \end{tabularx}
    \endgroup
\end{table}

\section{Additional Ablation Studies}

\subsection{Context Capacity and Temporal Coverage}
\begin{figure*}[t]
    \centering
    \includegraphics[width=\textwidth]{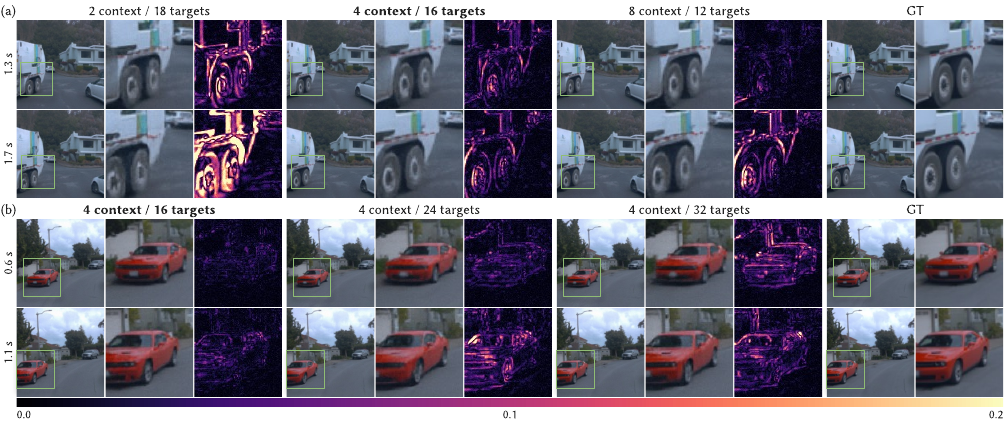}
    \caption{\textbf{Effect of context capacity and temporal coverage on Waymo.} (a) Within a 20-frame clip, 2, 4, or 8 frames serve as context and the remaining frames are targets. (b) Four context views span progressively longer clips with 16, 24, or 32 target views. Each row shows a target timestamp. For each setting, the full prediction is followed by a crop of the green-boxed region and its error map on a shared scale. Ground truth is shown at the right.}
    \label{fig:supp_ablation_spacing}
\end{figure*}

To complement the Waymo results in Table~4 of the main paper, we visualize how the number and spacing of context views affect target-view synthesis with a fixed-size scene latent.
As shown in Figure~\ref{fig:supp_ablation_spacing}, increasing the number of context views from two to four reduces errors around moving objects. These errors become more pronounced when the four context views are spread over a longer interval.

\subsection{Positional Encoding in the Scene Encoder}
Table~5 of the main paper compares 2D-RoPE, which encodes patch coordinates, with TP-RoPE, which also encodes camera geometry and capture time.
To isolate the contribution of capture time, we add PRoPE~\cite{li2025prope}, which incorporates the same projective camera encoding as TP-RoPE.
As shown in Table~\ref{tab:supp_rope_ablation}, encoding camera geometry improves full-image quality, and adding capture time yields further gains across the full image and in dynamic regions.

\begin{table}[t]
    \centering
    \caption{\textbf{Effect of scene encoder positional encoding on Waymo.}
    All variants use four context views and 16 target views, with the same training setup and the same target pose and time conditioning in the decoder.}
    \label{tab:supp_rope_ablation}
    \begingroup
    \footnotesize
    \setlength{\tabcolsep}{1pt}
    \renewcommand{\arraystretch}{1.12}
    \begin{tabularx}{\linewidth}{>{\raggedright\arraybackslash}p{0.25\linewidth}*{6}{>{\centering\arraybackslash}X}}
        \toprule
        \multirow{2}{*}{Encoding}
        & \multicolumn{3}{c}{Full Image}
        & \multicolumn{3}{c}{Dynamic Only} \\
        \cmidrule(lr){2-4}\cmidrule(lr){5-7}
        & \mbox{PSNR$\uparrow$}
        & \mbox{SSIM$\uparrow$}
        & \mbox{LPIPS$\downarrow$}
        & \mbox{PSNR$\uparrow$}
        & \mbox{SSIM$\uparrow$}
        & \mbox{LPIPS$\downarrow$} \\
        \midrule
        2D-RoPE & 30.99 & 0.826 & 0.289 & 27.08 & 0.785 & 0.393 \\
        PRoPE & 31.17 & 0.873 & 0.256 & 27.70 & 0.738 & 0.302 \\
        \midrule
        \rowcolor{gray!12}
        \textbf{TP-RoPE}
        & \textbf{32.91} & \textbf{0.888} & \textbf{0.229} & \textbf{28.77} & \textbf{0.838} & \textbf{0.162} \\
        \bottomrule
    \end{tabularx}
    \endgroup
\end{table}

\subsection{Scene Encoder and Supervision}
\begin{figure*}[p]
    \centering
    \includegraphics[width=\textwidth,height=0.62\textheight,keepaspectratio]{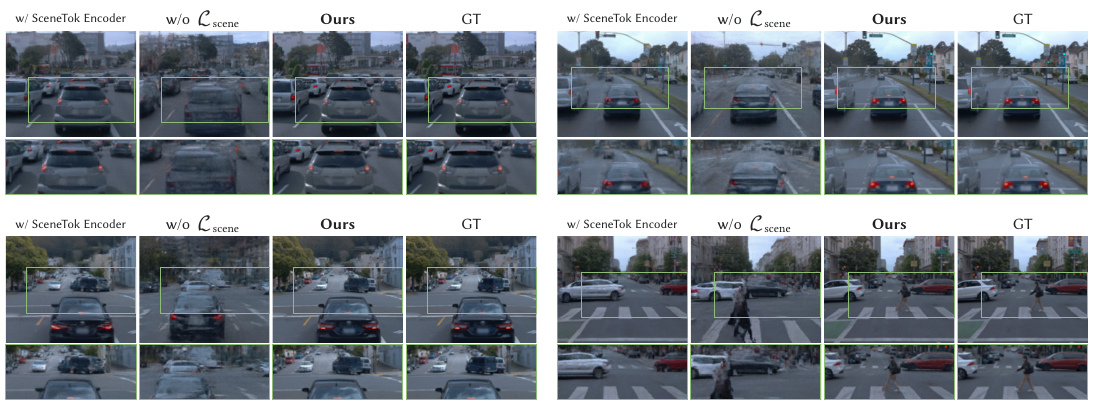}
    \caption{\textbf{Additional qualitative ablation results on Waymo.} Columns show the SceneTok encoder variant, the variant without $\mathcal{L}_{\mathrm{scene}}$, our full model, and ground truth. Each example presents full images and enlarged crops of the green-boxed regions.}
    \label{fig:supp_arch_ablation}
\end{figure*}

Figure~\ref{fig:supp_arch_ablation} provides additional qualitative results for the architecture and supervision ablations. Replacing our scene encoder with the scaled SceneTok encoder~\cite{asim2026scenetok} reduces dynamic object detail, while removing $\mathcal{L}_{\mathrm{scene}}$ introduces structural artifacts. Our full model better preserves object shape and appearance.

\section{Additional Comparison Results}
\label{sec:supp_baseline_comparisons}

\subsection{Additional Baselines on Waymo}
We extend the Waymo comparison to RoGe~\cite{lang2026roge} and TokenGS~\cite{ren2026tokengs} using the same four context views and 16 target views per sequence as in the main paper. Table~\ref{tab:supp_waymo_baselines} reports PSNR and SSIM over the full image and dynamic regions. The experimental results still support the accurate modeling of dynamic objects by our method.
Despite improvements in reconstruction accuracy, the main-paper results show that \method does not consistently achieve the lowest LPIPS. Our pixel-space denoising objective does not explicitly optimize perceptual feature similarity, which may partly explain this gap.

\begin{table}[t]
    \centering
    \caption{\textbf{Additional comparisons on Waymo.}}
    \label{tab:supp_waymo_baselines}
    \begingroup
    \footnotesize
    \setlength{\tabcolsep}{2pt}
    \renewcommand{\arraystretch}{1.12}
    \begin{tabularx}{\linewidth}{@{}lc*{4}{>{\centering\arraybackslash}X}@{}}
        \toprule
        \multirow{2}{*}{Method}
        & \multirow{2}{*}{\shortstack{Repr. Size\\(\#Floats)}}
        & \multicolumn{2}{c}{Full Image}
        & \multicolumn{2}{c}{Dynamic Only} \\
        \cmidrule(lr){3-4}\cmidrule(lr){5-6}
        & & PSNR$\uparrow$ & SSIM$\uparrow$
          & PSNR$\uparrow$ & SSIM$\uparrow$ \\
        \midrule
        RoGe~\cite{lang2026roge}
        & 2.39M & 28.66 & 0.862 & 23.61 & 0.660 \\
        TokenGS~\cite{ren2026tokengs}
        & 3.67M & 32.23 & \textbf{0.933} & 25.27 & 0.724 \\
        \midrule
        \rowcolor{gray!12}
        \textbf{\method}
        & 1.05M & \textbf{32.91} & 0.888 & \textbf{28.77} & \textbf{0.838} \\
        \bottomrule
    \end{tabularx}
    \endgroup
\end{table}

\subsection{Additional Qualitative Comparisons}
\begin{figure*}
    \centering
    \includegraphics[width=\textwidth]{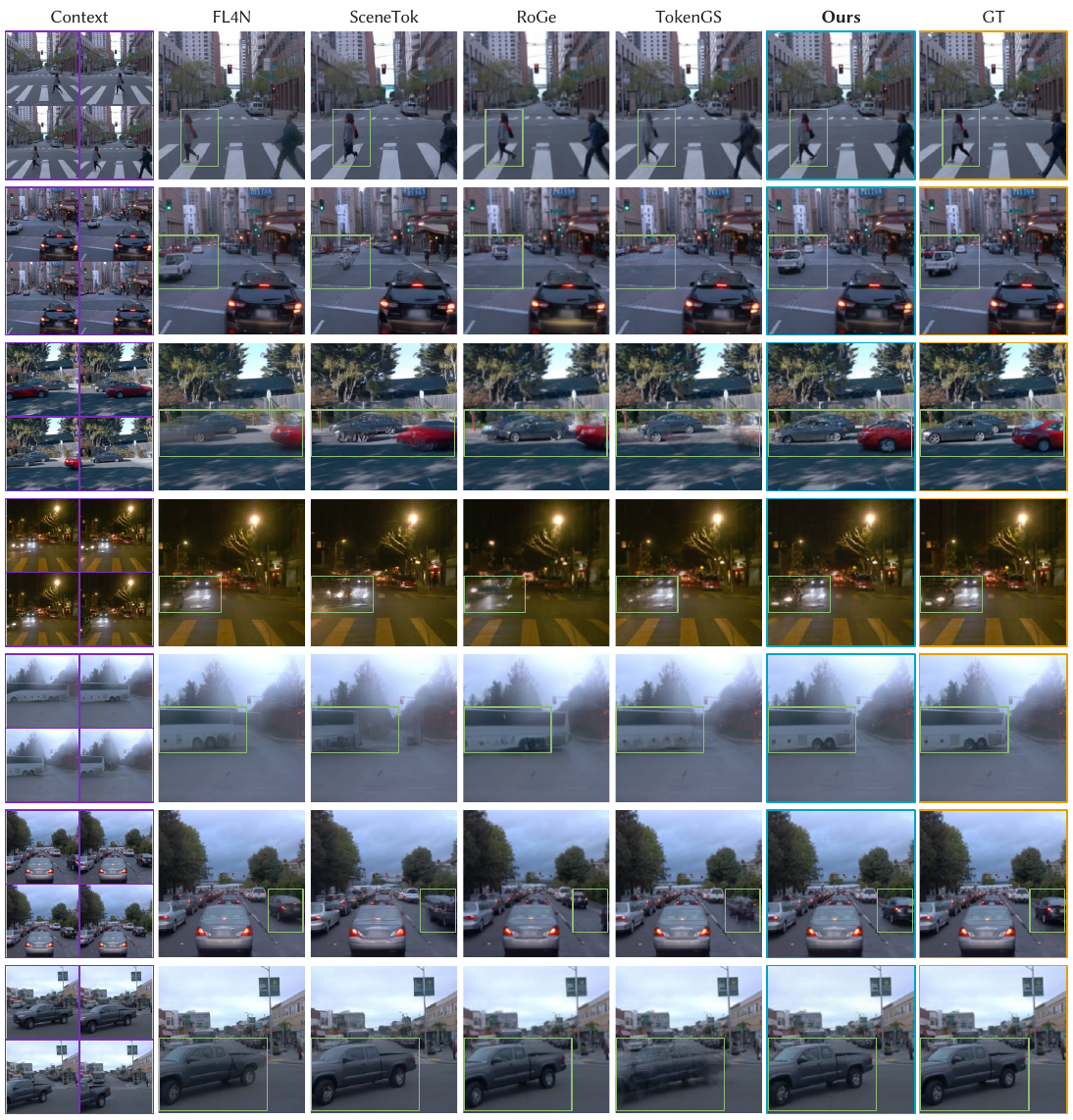}
    \caption{\textbf{Additional qualitative NVS comparisons on Waymo.} Each row shows four context views and one target view in the 4-context, 16-target setting. (\colorboxmark{contextpurple}, \colorboxmark{predcyan}, \colorboxmark{gtyellow}) denote the context views, our predictions, and ground-truth target views, respectively. Green boxes highlight moving objects. Across daytime, nighttime, and foggy scenes, \method better preserves dynamic object structure and appearance using a compact scene representation.}
    \label{fig:waymo_results_supp}
\end{figure*}

\begin{figure*}
    \centering
    \includegraphics[width=0.95\textwidth]{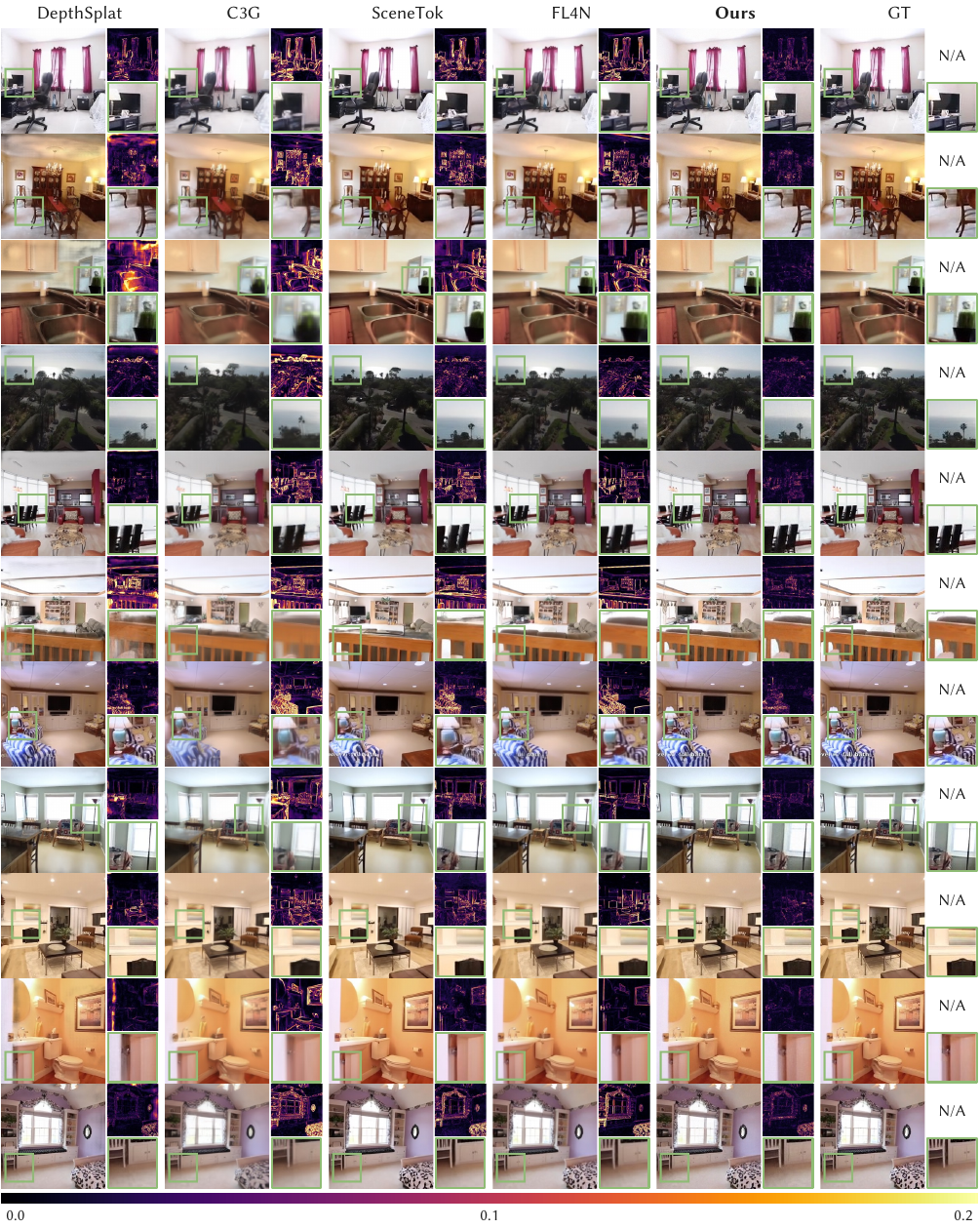}
    \caption{\textbf{Additional qualitative NVS comparisons on RE10K.} Each row shows one held-out view from a different scene in the 12-context, 128-target setting. Beside each prediction, the upper inset shows the mean absolute RGB error relative to the ground truth, and the lower inset magnifies the region outlined in green.}
    \label{fig:re10k_results_supp}
\end{figure*}

We show additional examples from Waymo and RE10K using the same baselines as in the main paper for each dataset. Figure~\ref{fig:waymo_results_supp} shows target views from daytime, nighttime, and foggy driving scenes on Waymo. Figure~\ref{fig:re10k_results_supp} shows held-out views from additional RE10K scenes, with error maps and enlarged crops for comparison.

\section{Additional Visualizations}

\subsection{Point Cloud Visualization}
\begin{figure*}[p]
    \centering
    \includegraphics[width=0.95\textwidth]{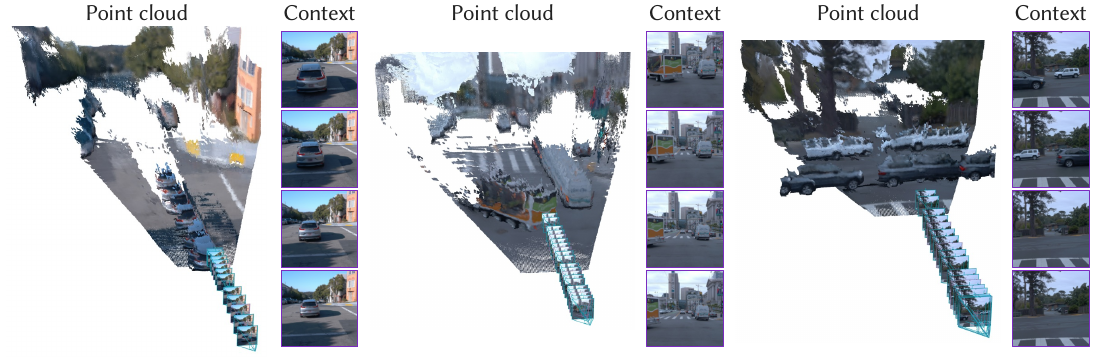}
    \caption{\textbf{Point clouds from predicted RGBD views on Waymo.} Each example shows the reconstructed point cloud on the left and the four context images on the right. Cyan frusta mark the target camera poses. In the rightmost scene, the car appears at successive positions as it crosses the intersection.}
    \label{fig:supp_pointcloud}
\end{figure*}

From four context images, \method encodes a single set of scene tokens and uses it to predict RGB and depth at each target camera pose and timestamp. To inspect the geometry of these predictions across views, we back-project the depth maps into a common world coordinate system and color the points with the predicted RGB values.
As shown in Figure~\ref{fig:supp_pointcloud}, the combined views preserve the street layout and the coarse shape of moving vehicles. The crosswalk stripes remain distinct on the road surface in the middle scene, and the outline of the passing car is recognizable at successive positions in the rightmost scene. 

\subsection{Rendered Sequences}
\begin{figure*}[p]
    \centering
    \includegraphics[width=\textwidth,height=\textheight,keepaspectratio]{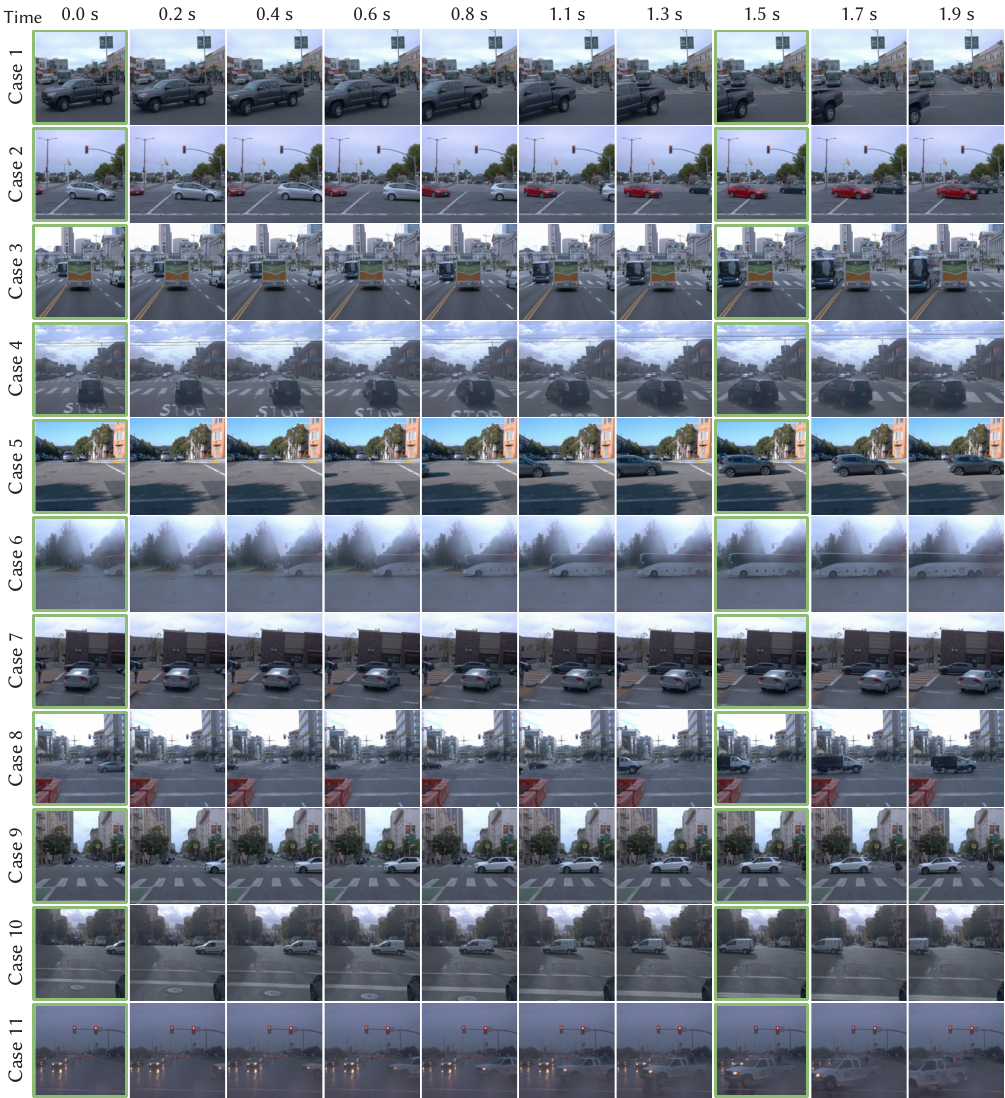}
    \caption{\textbf{Novel view synthesis over time on Waymo.} Each row shows one driving sequence. Green borders mark the first and last of four context frames (0.0 and 1.5\,s). The eight unbordered frames show predictions, including extrapolations at 1.7 and 1.9\,s.}
    \label{fig:supp_seq_waymo}
\end{figure*}

\begin{figure*}[p]
    \centering
    \includegraphics[width=\textwidth,height=\textheight,keepaspectratio]{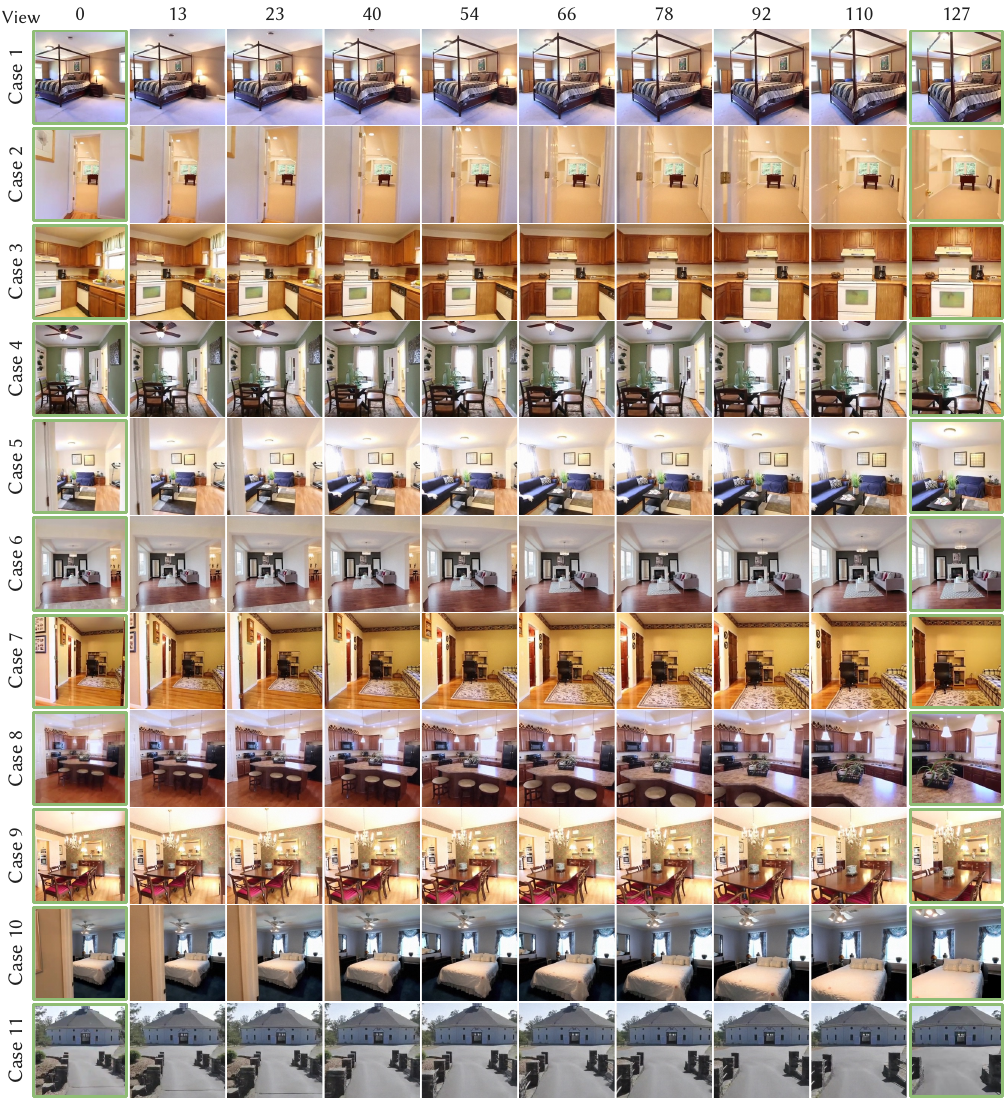}
    \caption{\textbf{Novel view synthesis along RE10K camera trajectories.} Each row shows ten views from a 128-view sequence. Green borders mark the first and last of twelve context views. The eight columns between them show predictions at the labeled held-out viewpoints.}
    \label{fig:supp_seq_re10k}
\end{figure*}

On Waymo and RE10K, we render each sequence from a fixed set of context images and show selected target views along the camera trajectory.
As shown in Figures~\ref{fig:supp_seq_waymo} and~\ref{fig:supp_seq_re10k}, the Waymo renderings capture changes in traffic over time, while the RE10K renderings show fine scene details from different viewpoints.
Please see our \textit{supplementary video} for more long-sequence results.

\end{document}